\documentclass[runningheads]{llncs}
\usepackage{graphicx}

\usepackage{tikz}
\usepackage{comment} 
\usepackage{amsmath,amssymb} 
\usepackage{color}

\usepackage{booktabs}
\usepackage{multirow}
\usepackage{url}
\usepackage[linesnumbered,boxed,ruled,commentsnumbered]{algorithm2e}
\usepackage{pifont}
\usepackage{marvosym}
\usepackage{colortbl}
\definecolor{gray}{rgb}{0.85,0.85,0.85}

\begin{document}
\pagestyle{headings}
\mainmatter
\def\ECCVSubNumber{6474}  

\title{Learning to Learn Parameterized Classification Networks for Scalable Input Images} 

\titlerunning{SAN: Scale Adaptive Network for Scalable Input Images}
%
\author{Duo Li\inst{1,2}\thanks{indicates intern at Intel Labs China. \textsuperscript{\Letter} indicates corresponding authors.} \and
	Anbang Yao\inst{2}\textsuperscript{\Letter} \and
	Qifeng Chen\inst{1}\textsuperscript{\Letter}}
\authorrunning{D. Li, A. Yao and Q. Chen}
%
\institute{The Hong Kong University of Science and Technology \and
	Intel Labs China \\
	\email{duo.li@connect.ust.hk}\qquad
	\email{anbang.yao@intel.com}\qquad
	\email{cqf@ust.hk}}
\maketitle

\begin{abstract}
	Convolutional Neural Networks (CNNs) do not have a predictable recognition behavior with respect to the input resolution change. This prevents the feasibility of deployment on different input image resolutions for a specific model. To achieve efficient and flexible image classification at runtime, we employ meta learners to generate convolutional weights of main networks for various input scales and maintain privatized Batch Normalization layers per scale. For improved training performance, we further utilize knowledge distillation on the fly over model predictions based on different input resolutions. The learned meta network could dynamically parameterize main networks to act on input images of arbitrary size with consistently better accuracy compared to individually trained models. Extensive experiments on the ImageNet demonstrate that our method achieves an improved accuracy-efficiency trade-off during the adaptive inference process. By switching executable input resolutions, our method could satisfy the requirement of fast adaption in different resource-constrained environments. Code and models are available at \url{https://github.com/d-li14/SAN}.
\end{abstract}

\section{Introduction}

Although CNNs have demonstrated their dominant power in a wide array of computer vision tasks, their accuracy does not scale up and down with respect to the corresponding input resolution change. Typically, modern CNNs are constructed by stacking convolutional modules in the body, a Global Average Pooling (GAP) layer and a Fully-Connected (FC) layer in the head. When input images with different sizes are fed to a CNN model, the convolutional feature maps also vary in their size accordingly, but the subsequent GAP operation could reduce all the incoming features into a tensor with $1 \times 1$ spatial size and equal amount of channels. Thanks to the GAP layer, even trained on specific-sized input images, modern CNNs are also amenable to processing images of other sizes during the inference phase.
\begin{figure}[htbp]
	\vskip -0.1in
	\begin{center}
		\includegraphics[width=.9\linewidth]{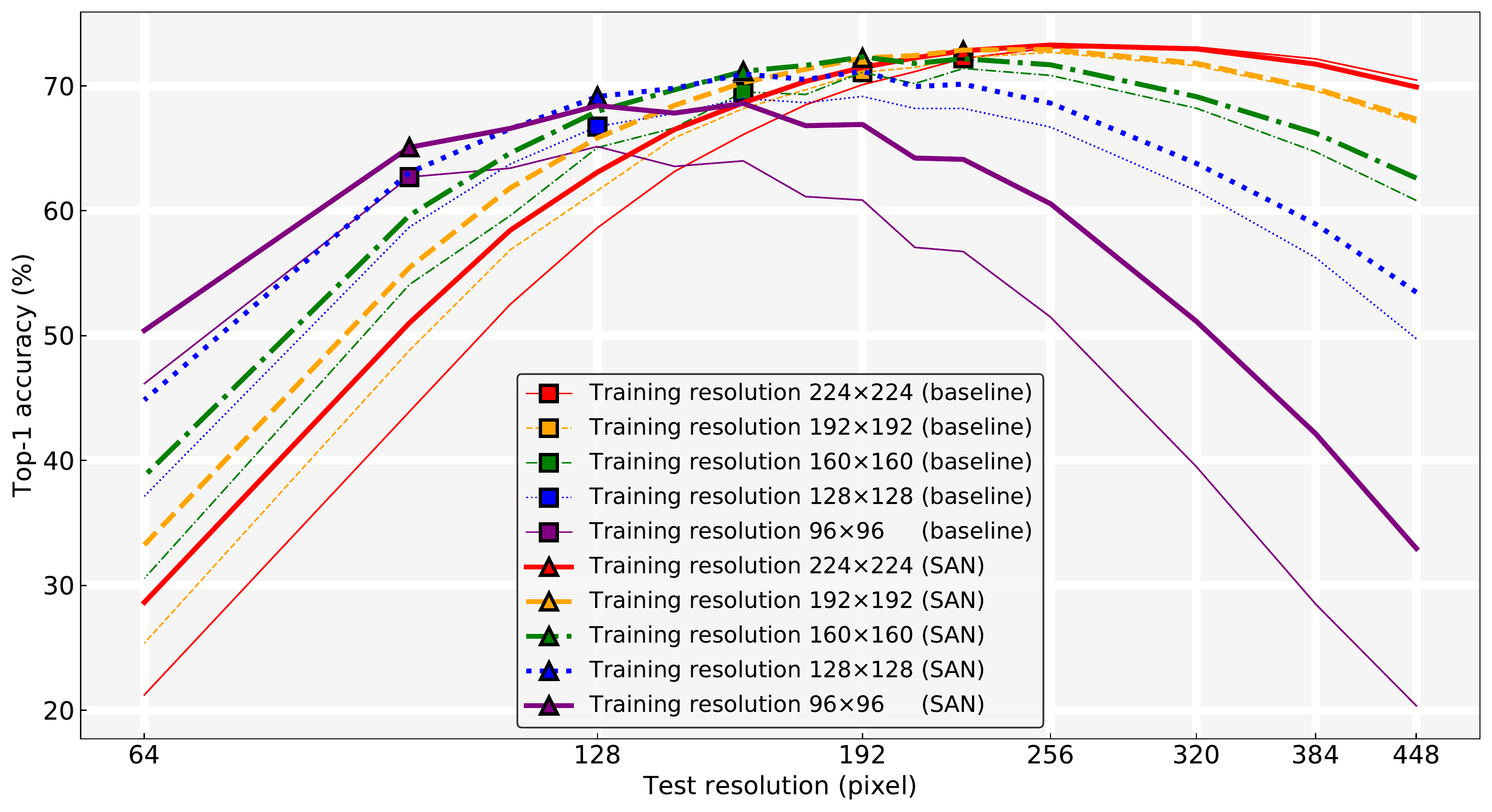}
	\end{center}
	\vskip -0.3in
	\caption{Validation accuracy envelope of our proposed SAN with MobileNetV2 on ImageNet. Curves with the same color/style represent the results of models trained with the same input resolutions. The x-axis is plotted in the logarithmic scale. $\blacktriangle$ and $\blacksquare$ indicate the spots when test resolution meets the training one.}
	\label{fig:envelop}
	\vskip -0.25in
\end{figure}
However, the primary concern lies in that their performance is vulnerable to \textit{scale deviation}\footnote{The concept of \textit{scale deviation} will be discussed detailedly in Section~\ref{sec:related}.}, exhibiting severe deterioration when evaluating images of varying sizes at the inference time, as illustrated in Fig.~\ref{fig:envelop}. Therefore, as done in a series of efficient network designs \cite{2017arXiv170404861H}\cite{Sandler_2018_CVPR}\cite{Howard_2019_ICCV}\cite{Zhang_2018_CVPR}\cite{Ma_2018_ECCV}\cite{hbonet}, in order to adapt to real-time computational demand from the aspect of input resolution, it is necessary to train a spectrum of models from scratch using input images of different resolutions. These pre-trained models are then put into a storage pool and individually reserved for future usage. Client-side devices have to retrieve pertinent models based on requirements and available resources at runtime, which will largely impede the flexibility of deployment due to inevitable downloading and offloading overheads. To flexibly handle the real-time demand on various resource-constrained application platforms, a question arises: under the premise of not sacrificing or even improving accuracy, is it possible to learn a controller to dynamically parameterize different visual classification models having a shared CNN architecture conditioned on the input resolutions at runtime?

To echo the question above, a scale-adaptive behavior is anticipated for the controller to acquire. That implies, given a CNN structure, to instantiate different main networks for each certain image resolution, the controller should have prior knowledge about switching between scaled data distributions and tactfully bridging the scaling coefficients with network parameters. We propose that appropriate data scheduling and network layer manipulations could lend this attribute to the controller. Specifically, we synthesize image patches with a set of training resolutions and employ meta networks as the controller to integrate diverse knowledge from these varying resolutions. The meta learners redistribute the gathered scale knowledge by generating convolutional kernels for multiple main networks conditioned on their assigned input resolutions respectively. Due to the tight relationship between Batch Normalization (BN)~\cite{pmlr-v37-ioffe15} layers and scaled activation distributions, all parameters and statistics of BN layers in each main network are maintained in a privatized manner. The main networks are collaboratively optimized following a mixed-scale training paradigm. The meta networks are optimized via collecting gradients passed through different main networks, such that information from multi-scale training flows towards the controller. Furthermore, aiming at leveraging the knowledge extracted with large resolutions to advance the performance on small ones, a scale distillation technique is utilized on the fly via taking the probabilistic prediction based on large resolutions as smoothed supervision signals. These aforementioned ingredients are coherently aggregated, leading to our proposed Scale Adaptive Network (SAN) framework, which is scalable by design and also generally applicable to prevailing CNN architectures.

During inference, given an input image, the learned controller could parameterize the visual classification model according to its resolution, showing consistently improved accuracy compared with the model individually trained on the corresponding resolution, as demonstrated in Fig.~\ref{fig:envelop}. Therefore, client devices almost merely need to reserve a single meta network and the computation graph of the backbone architecture whose parameters could be dynamically generated based on the evaluation resolution. Depending on the ability to flexibly switch target resolutions, the inference process is controllable to meet the on-device latency requirements and the accuracy expectation. Furthermore, the generated models could adapt smoothly to a wide range of input image sizes, even under the circumstances with severe problems of scale deviation. Benefiting from in-place parameterization and performance resilience on a spectrum of evaluated resolutions for each individual model, the one-model-fits-all style addresses the major obstacle of application to various scenarios.

Our main contributions can be summarized as follows:
\begin{description}
	\item[\ding{111}] We employ meta networks to decide the main network parameters conditioned on its input resolution at runtime. Little research attention has been paid to this kind of meta-level information before us. We also extend the scope of knowledge distillation, based on the same image instance with different resolutions, which is also a rarely explored data-driven application scenario.
	\item[\ding{111}] We develop a new perspective of efficient network	design by combining aforementioned two components to permit adaptive inference on universally scaled input images, make a step forward in pushing the boundary of the resolution-accuracy spectrum and facilitating flexible deployment of visual classification models among switchable input resolutions.
\end{description}

\section{Related Work}
\label{sec:related}

We briefly summarize related methodologies in previous literature and analyze their relations and differences with our approach as follows.

\textbf{Scale Deviation.} FixRes~\cite{touvron2019FixRes} sheds light on the distribution shift between the train and test data, and quantitatively analyzes its effect on apparent object sizes and activation statistics, which arises from inconsistent data pre-processing protocols during training and testing time. The discrepancy between train and test resolution is defined as \textit{scale deviation} in this work. We would like to clarify that the issue of scale deviation exists across universally scaled images in the visual classification task, which shows a clear distinction compared to another popular phenomenon named \textit{scale variance}. The problem of \textit{scale variance} is more commonly identified among instances of different sizes within a single image, especially in the images of the MS COCO~\cite{10.1007/978-3-319-10602-1_48} benchmark for object detection. Typically hierarchical or parallel multi-scale feature fusion approaches~\cite{Szegedy_2016_CVPR,Lin_2017_CVPR,Zhou_2018_CVPR,Li_2019_ICCV,Zhao_2017_CVPR,Chen_2018_ECCV} are utilized to address this problem, which has a loose connection with our research focus nevertheless. To handle the initial problem of \textit{scale deviation}, FixRes proposes a simple yet effective strategy that prefers increased test crop size and fine-tunes the pre-trained model on the training data with the test resolution as a post-facto compensation. Notably, FixRes lays emphasis on calibrating BN statistics over the training data modulated by the test-size distribution, which is of vital importance to remedy the activation distribution shift. In comparison, we could use a proxy or data-free inference method to avoid the calibration of BN statistics, thus no post-processing steps are involved after end-to-end network training. We further provide empirical comparison in Section~\ref{sec:experiments}. Specializing BN layers for network adaption between a few different tasks has been adopted in domain adaption~\cite{NIPS2017_6654}, transfer learning~\cite{mudrakarta2018k}, adversarial training~\cite{Xie_2020_CVPR} and optimization of sub-models in a super net~\cite{yu2018slimmable}\cite{Yu_2019_ICCV}. Inspired by them, we also overcome the shortcoming of statistic discrepancy through privatizing statistics and trainable parameters in the BN layers of each main network.

\textbf{Meta Learning.} Meta learning, or learning to learn, has come into prominence progressively in the field of artificial intelligence, which is supposed to be the learning mechanism in a more human-like manner. The target of meta learning is to advance the learning procedure at two coupled levels, enabling a \textit{meta learning} algorithm to adapt to unseen tasks efficiently via  utilizing another \textit{base learning} algorithm to extract transferable prior knowledge within a set of auxiliary tasks. The hypernetwork~\cite{Ha_2017_ICLR} is proposed to efficiently generate the weights of the large main network using a small auxiliary network, in a relaxed form of weight-sharing across layers. Bertinetto \textit{et al.}~\cite{NIPS2016_6068} concentrate on the one-shot learning problem and uses a \textit{learnet} to map a single training exemplar to weights of the main network in one go. Andrychowicz \textit{et al.}~\cite{NIPS2016_6461} replace the standard optimization algorithm with an LSTM optimizer to generate an update for the main network in a self-adaptive manner. Meta-LSTM~\cite{Ravi_2017_ICLR} further develops this idea by revealing the resemblance between cell states in an LSTM and network weights in a CNN model with respect to their update process. SMASH~\cite{brock2018smash} applies the HyperNet to Neural Architecture Search (NAS) by transforming a binary encoding of the candidate network architecture to its corresponding weights. In our framework, meta networks, or to say hypernetworks, are responsible for generating weights of convolutional layers in the main network according to scale-related meta knowledge of the input data. 

\textbf{Knowledge Distillation.} Knowledge Distillation (KD) is based on a teacher-student knowledge transfer framework in which the optimization of a lower-capacity student neural network is guided by imitating the soft targets~\cite{2015arXiv150302531H}\cite{Zhang_2018_CVPR_DML} or intermediate representation~\cite{romero2015fitnets}\cite{Yim_2017_CVPR}\cite{Zagoruyko2017AT} from a large and powerful pre-trained teacher model. Inspired by curriculum learning~\cite{Bengio:2009:CL:1553374.1553380}, RCO~\cite{Jin_2019_ICCV} promotes the student network to mimic the entire route sequence that the teacher network passed by. RKD~\cite{Park_2019_CVPR} and CCKD~\cite{Peng_2019_ICCV} both exploit structural relation knowledge hidden in the embedding space to achieve correlation congruence. Several recent works also demonstrate the effectiveness of KD in improving the teacher model itself by self-distillation~\cite{2018arXiv180502641B}\cite{pmlr-v80-furlanello18a}. We introduce the scale distillation into our framework to transfer the scale and structural information of the same object instances from large input images to smaller ones at each step throughout the whole training procedure. The knowledge distillation process emerges among different main networks with the same network structure that travel along the same route sequence of optimization, without the assistance of external teacher models. 

\section{Approach}

\begin{figure}[htbp]
	\vskip -0.2in
	\begin{center}
		\includegraphics[width=.85\linewidth,trim=10 20 20 20,clip]{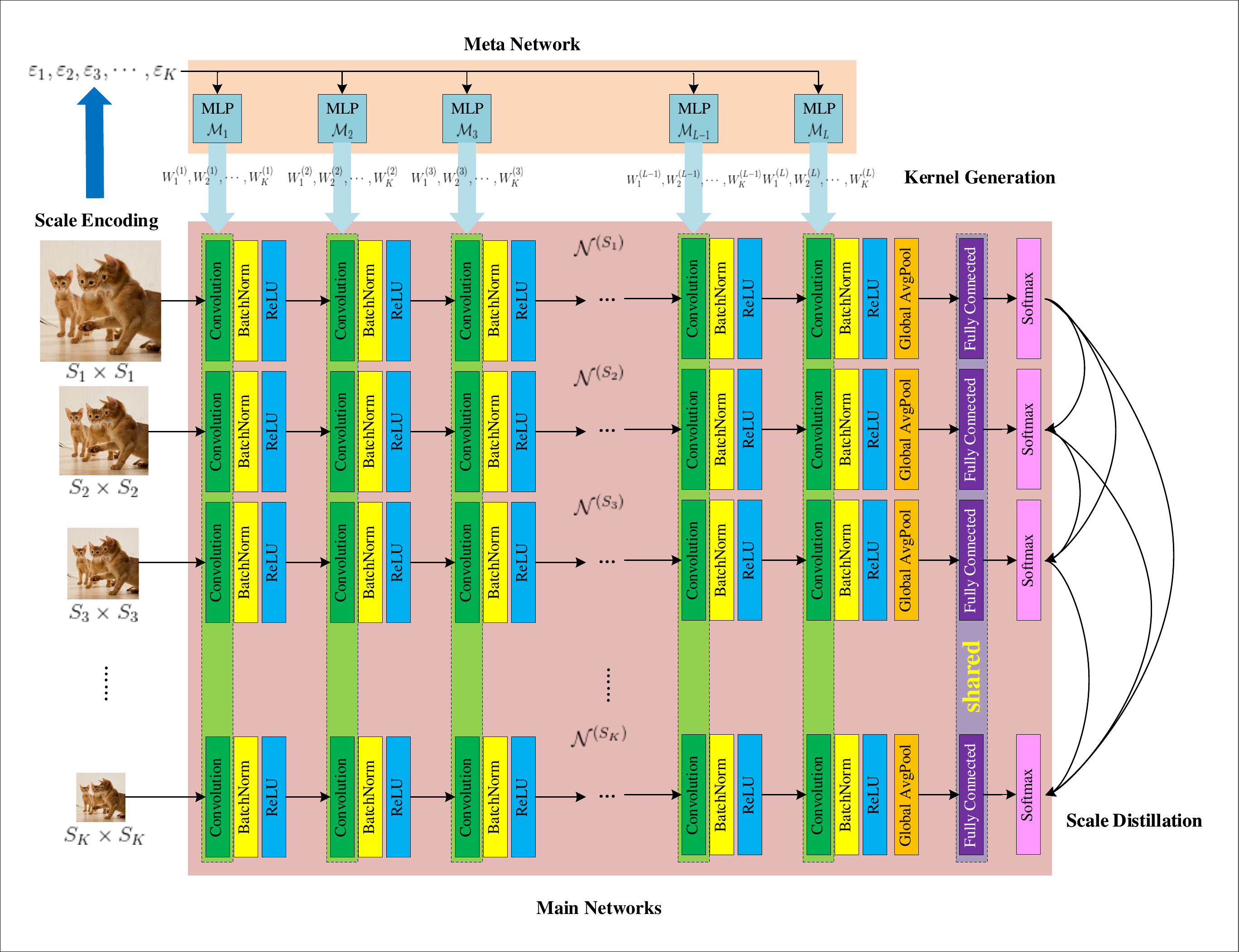}
	\end{center}
	\vskip -0.3in
	\caption{Schematic illustration of our proposed Scale Adaptive Network framework. The input image scales $S_1, S_2, \cdots, S_K$ are linearly transformed into a set of encoding scalars $\varepsilon_1, \varepsilon_2,  \cdots, \varepsilon_K$, which are fed into the MLP meta network $\mathcal{M}_l, l=1,2,\cdots,L$, generating weights $\{W_1^{(l)}, W_2^{(l)}, \cdots, W_K^{(l)}\}$ of the \textit{l}\textsuperscript{th} convolutional layer for each main network $\mathcal{N}^{(S_i)}$ associated with one certain input scale $S_i, i=1,2,\cdots,K$. BN layers are privatized and FC layers are shared among different main networks. The Scale Distillation process is performed in a top-down fashion. Best viewed in electronic version.}
	\label{fig:teaser}
	\vskip -0.3in
\end{figure}

A schematic overview of our method is presented in Fig.~\ref{fig:teaser}. The key innovations lie in the employment of meta networks in the body and scale distillation at the end. In this section, we elaborate on the insights and formulations of them.

\subsection{Network Architecture}

Motivated by the target of achieving the scale-adaptive peculiarity of CNNs, we speculate that scale-invariant knowledge should be exploited regarding each input image in its heterogeneous modes of scales. Although the same object instances pose significant scale variation in these modes, they could share common structural relations and geometric characteristics. A meta learner built upon the top of CNNs is expected to extract and analyze this complementary knowledge across scales. Then it injects this prior knowledge into underlying CNNs through parameterization, making them quickly adapt to a wide range of new tasks of visual classification which may include unseen scale transformations.

In this spirit, we denote the model optimized on training images with the resolution $S \times S$ as $\mathcal{N}^{(S)}$ and consider a group of main networks optimized on a set of input resolutions $\mathbb{S}=\{S_1, S_2, \cdots, S_k\}$ respectively. We also construct a cohort of meta networks $\mathcal{M}=\{\mathcal{M}_1, \mathcal{M}_2, \cdots, \mathcal{M}_L\}$ to generate convolutional kernels conditioned on the scale $S$ of input images for an $L$-layer convolutional neural network $\mathcal{N}^{(S)}$. As illustrated in Fig.~\ref{fig:teaser}, bidirectional information flow is established between the shared meta network and each individual main network by the means of \textbf{Scale Encoding} and \textbf{Kernel Generation} respectively. All meta networks are instantiated by the same network structure of a Multi-Layer Perceptron (MLP). 

\textbf{Scale Encoding.} The input to each MLP is an encoding scalar that contains information about the relative scale of training examples. Since the down-sampling rate of prevalent CNNs is $1/32$, we heuristically apply a normalization policy to linearly transform the input scale $S$ to an encoding $\varepsilon=0.1 \times S/32$. 

\textbf{Kernel Generation.} The output dimension of an MLP equals the dimension of its corresponding convolutional kernel but in the flattened form. For example, a 1-in, $(C_{out}C_{in}K^2)$-out MLP is built to adaptively generate $C_{out}$ groups of convolutional kernels, each group containing $C_{in} $ kernels with the same size of $K \times K$. Compared with hypernetworks~\cite{Ha_2017_ICLR} that map each learnable embedding vector to the weights of its corresponding layer with one auxiliary network, our mechanism assigns each meta learner to generate weights for the corresponding layer with one common input encoding scalar regarding a specific main network.

With these well-defined meta learners, parameters of convolutional layers in the main network $\mathcal{N}^{(S)}$ can be generated and associated with a specific input scale $S$. As emphasized by previous works~\cite{touvron2019FixRes}, BN layers should be delicately handled for the sake of inconsistency between data distributions of varying scales. Note that parameters in the BN layers usually occupy a negligible portion of the parameters within the whole model, \textit{e}.\textit{g}., less than 1\% of the total amount in most cases, we opt for a straightforward yet effective strategy by maintaining individual BN layers, denoted as $\text{BN}^{(S)}$, for each scale-specified main network $\mathcal{N}^{(S)}$. Injecting these conditional normalization layers can lend more inherent adaptability and scalability to intermediate feature representations to accommodate scale discriminative knowledge. By contrast, with regard to the last FC layer that occupies a considerable amount of parameters, a shared one is designated to fit any potential main networks.

To take full advantage of various sources of scale information, we propose a mixed-scale optimization scheme accordingly. With the label-preserved transformation, each image in the training set $\mathcal{D}=\{(\boldsymbol{x}_1, y_1), (\boldsymbol{x}_2, y_2), \cdots, (\boldsymbol{x}_N, y_N)\}$ is resized to a series of scales, \textit{e}.\textit{g}.,  $S_1, S_2, \cdots, S_k$, where $y_j, j=1,2,\cdots,N$ is the ground truth category label of the sampled image $\boldsymbol{x}_j$. For a certain scale $S_i$, we encode it as $\varepsilon^{(S_i)}$ and map the encoding scalar to a fully parameterized main network $\mathcal{N}^{(S_i)}$ through the meta learners. Transformed samples with the same size $S_i$ are assembled to form a resized version of the original training set denoted as $\mathcal{D}^{(S_i)}$ and fed into the corresponding main network $\mathcal{N}^{(S_i)}$. For each pair of $\mathcal{D}^{(S_i)}$ and $\mathcal{N}^{(S_i)}$, it follows the standard optimization procedure of convolutional neural networks via minimizing the cross-entropy loss. Since our objective is to optimize the overall accuracy under different settings of scales, no scale is privileged and the total classification loss is an un-weighted sum of the individual losses, represented as
\begin{equation}
\mathcal{L}_{\text{CE}} = \sum_{i=1}^{k}\sum_{j=1}^{N}\mathcal{L}_{\text{CE}}(\boldsymbol{\theta}^{(S_i)}; \boldsymbol{x}_j^{(S_i)}, y_j),
\end{equation}
where $k$ is the number of resize transformations and $\boldsymbol{\theta}^{(S_i)} = \{\mathcal{M}_l(\varepsilon^{(S_i)})\ \vert\ l=1,2,\cdots,L\}\cup\{\text{BN}^{(S_i)}, \text{FC}\}$ are weights of the network $\mathcal{N}^{(S_i)}$ where convolutional layers are directly generated by meta learners. The parameters in meta networks $\mathcal{M}$ are optimized simultaneously following the chain rule as the aforementioned weight generation operations are completely differentiable.

The number of hidden layers and units in the MLP could be tuned. In our main experiments, the meta network is chosen as a single layer MLP for the purpose of effectiveness and efficiency (validated by ablation experiments later). It could be represented as an FC layer with the weight $\textbf{W}_l$ and bias $\textbf{b}_l$
\begin{equation}
\label{eqn:fc}
\mathcal{M}_l(\varepsilon^{(S_i)}) = \varepsilon^{(S_i)}\textbf{W}_l + \textbf{b}_l.
\end{equation}
Due to the existence of the bias term, the output convolutional kernels are \textit{not} simply distinct up to a scaling factor $\varepsilon^{(S_i)}$ across different main networks. To be cautious, we also examine the value of $\textbf{W}_l$ and $\textbf{b}_l$ regarding each layer. They are of the same size and the same order of magnitude in almost all cases, indicating that the weights and biases have an equivalently important influence on the learning dynamics. The ratio of $\textbf{W}_l$ to $\textbf{b}_l$ in each layer is reported in the supplementary materials.

\subsection{Scale Distillation}

High-resolution representations~\cite{Sun_2019_CVPR} can contain finer local feature descriptions and more discernable semantic meaning than the lower-resolution ones, hence it is appropriate to utilize the probabilistic prediction over larger-scale inputs to offer auxiliary supervision signals for smaller ones, which will be referred to as \textit{Scale Distillation} in this context. Specifically, a Kullback-Leibler divergence $D_{\text{KL}}(\cdot\Vert\cdot)$ is calculated between each pair of output probability distributions among all main networks, which leads to an additional representation mimicking loss shaped in a top-down manner as follows
\begin{equation}
\mathcal{L}_{\text{SD}} = \sum_{i=1}^{k}\sum_{\substack{j=1\\S_j<S_i}}^{k} D_{\text{KL}}(\boldsymbol{p}^{(S_i)} \Vert \boldsymbol{p}^{(S_j)}),
\end{equation}
where $\boldsymbol{p}^{(S_i)}$ denotes the probabilistic distribution prediction with respect to object categories outputted by the main network $\mathcal{N}^{(S_i)}$.

Our mixed-scale training mechanism naturally supports scale distillation as an in-place operation. During each training step, we take the predicted labels of one model and fix them as the soft training labels for other models processing smaller input samples, which can be implemented on the fly without introducing further memory overheads in practice. Compared with conventional KD methodology~\cite{2015arXiv150302531H}, a main network in our framework may be both the teacher model and the student model, depending on its matched counterpart model. Furthermore, the cohort of main networks are of the same model size and optimized in a single-shot rather than two-stage manner.

Then, the overall optimization objective of our framework is to minimize the combined loss function
\begin{equation}
\mathcal{L} = \alpha \mathcal{L}_{\text{CE}} + \beta \mathcal{L}_{\text{SD}},
\end{equation}
where $\alpha$ and $\beta$ are positive weight coefficients to balance the cross-entropy loss and the scale distillation loss. In our main experiments, we do not make much investment in tuning these hyperparameters and find that simply setting $\alpha=\beta=1$ leads to satisfactory performance, which demonstrates the robustness of our proposed optimization scheme in some sense.

Intuitively condensing networks for different purposes into a shared framework tends to bring about performance degradation compared to individually trained ones, since they might exhibit inconsistent learning dynamics. However, we surmise that our performance improvements could originate from a relaxed form of knowledge transfer across different scales. According to Eqn.~\ref{eqn:fc}, weights and biases of the meta networks respectively enforce convolution parameter scaling and sharing across different main networks. The generated weights for one scale also depend on the information from any other different scales due to the joint training. By feat of the shared meta networks and a collaborative training regime, the knowledge interaction process between models may be interpreted as an \textit{implicit distillation}. In addition to the implicit information sharing mechanism above, we further develop an \textit{explicit distillation} technique to aggressively advance this knowledge transfer process, presented as Scale Distillation.

\subsection{Inference}
\label{sec:inference}
For inference, let the selected training resolution range be $\mathbb{S}=\{S_1, S_2, \cdots, S_k\}$ and the test resolution be $T$, we first search the nearest resolution $S(T) \in \mathbb{S}$ for $T$, then feed the scale encoding $\varepsilon^{(T)}$ to the pre-trained meta network to parameterize convolutional layers for the main network and the BN layers reserved for $S(T)$ during training could be applied directly. Finally the \textit{ideal inference} is realized by sending the test image to the network parameterized by
\begin{equation}
\label{eqn:ideal}
\boldsymbol{\theta}^{(T)}_{\text{ideal}} = \{\mathcal{M}_l(\varepsilon^{(T)}) \vert l=1,2,\cdots,L\}\cup\{\text{BN}^{(S(T))}, \text{FC}\}.
\end{equation}
If $S(T)$ is \textit{not} close enough to $T$ regarding its value, the trainable parameters of BN (scale and shift parameter) could still be applied directly but calibration of BN statistics (running mean and variance) is necessary for retaining a decent performance (will be shown later in the ablation studies). Intriguingly, we find that simply substituting the scale encoding $\varepsilon^{(T)}$ by $\varepsilon^{(S(T))}$ to match its corresponding BN layers is ready to make compensation. It says that the runtime network could also be dynamically parameterized by
\begin{equation}
\label{eqn:proxy}
\boldsymbol{\theta}^{(T)}_{\text{proxy}} = \{\mathcal{M}_l(\varepsilon^{(S(T))}) \vert l=1,2,\cdots,L\}\cup\{\text{BN}^{(S(T))}, \text{FC}\},
\end{equation}
to achieve \textit{approximate} performance as that above (recognition accuracy by using $\boldsymbol{\theta}^{(T)}_{\text{ideal}}$ and $\boldsymbol{\theta}^{(T)}_{\text{proxy}}$ for inference is comprehensively compared in the ablation experiments). Since using $\boldsymbol{\theta}^{(T)}_{\text{proxy}}$ for inference would be relatively convenient once the pre-trained main network weights $\mathcal{M}(\varepsilon^{(S(T))})$ are already on hand, we opt for this \textit{proxy inference} method as an alternative for performance benchmark throughout Section~\ref{sec:experiments}, if no further specification. In practice, clients could choose either option to achieve similar performance, depending on whether the desired convolutional weights are handily accessible or even on hand.

Summarily, the \textbf{ideal inference} in Eqn.~\ref{eqn:ideal} uses test-specified encoding and calibration for BN statistics (if necessary) while the \textbf{proxy inference} in Eqn.~\ref{eqn:proxy} uses training-specified encoding and no calibration. The concrete algorithms of optimization and inference are provided in the supplementary materials.

\section{Experiments}
\label{sec:experiments}

We present extensive experimental results on ImageNet using various prevailing CNN architectures and conduct controlled experiments for introspection.

\subsection{Main Results}
\label{sec:main}

Our method is evaluated on the large-scale ImageNet~\cite{imagenet_cvpr09} dataset, which is a very challenging image recognition benchmark including over 1.2 million training images and 50,000 validation images belonging to 1,000 object categories. We follow the standard practice for data augmentation~\cite{He_2016_CVPR}, utilizing random resizing and cropping operations together with horizontal flipping to generate image patches with the desired resolutions. During evaluation, we crop the center region from each transformed image of which the shorter side is resized to satisfy the crop ratio of 0.875. We report top-1 validation error in all the following tables.

\begin{table}[htbp]
	\caption{Comparison of ResNet-18, MobileNetV2 and ResNet-50 (\textit{from top to bottom}) baseline models (\textit{left panel}) and SAN (\textit{right panel}) on a spectrum of test resolutions.}
	\label{tab:san-comparison}
	\vskip -0.1in
	\resizebox{.5\linewidth}{!}{
		\begin{tabular}{c|ccccccccccc}
			Train $\backslash$ Test & 256 & 224 & 208 & 192 & 176 & 160 & 144 & 128 & 112 & 96 & 64 \\
			\midrule[0.1em]
			224 & \cellcolor{gray}71.90 & \cellcolor{gray}70.97 & 69.14 & 68.72 & 66.20 & 64.68 & 60.36 & 56.85 & 50.17 & 42.55 & 20.40 \\
			192 & 71.71 & 71.25 & \cellcolor{gray}69.64 & \cellcolor{gray}69.75 & 67.70 & 67.04 & 63.23 & 60.47 & 54.21 & 47.73 & 25.10 \\
			160 & 70.72 & 70.70 & 69.50 & 70.14 & \cellcolor{gray}68.47 & \cellcolor{gray}68.48 & 65.61 & 63.80 & 58.54 & 53.78 & 31.03 \\
			128 & 67.41 & 68.51 & 67.83 & 69.06 & 67.81 & 68.56 & \cellcolor{gray}66.51 & \cellcolor{gray}66.36 & 61.88 & 58.29 & 38.58 \\
			96 & 54.68 & 58.87 & 58.46 & 62.39 & 61.62 & 64.84 & 63.42 & 65.51 & \cellcolor{gray}62.39 & \cellcolor{gray}62.56 & \cellcolor{gray}47.48 \\
		\end{tabular}
	}
	\resizebox{.5\linewidth}{!}{
		\begin{tabular}{c|ccccccccccc}
			Train $\backslash$ Test & 256 & 224 & 208 & 192 & 176 & 160 & 144 & 128 & 112 & 96 & 64 \\
			\midrule[0.1em]
			224 & \cellcolor{gray}72.96 & \cellcolor{gray}72.65 & 71.89 & 71.89 & 70.39 & 69.62 & 67.15 & 65.36 & 59.96 & 55.57 & 34.98 \\
			192 & 72.60 & 72.79 & \cellcolor{gray}72.10 & \cellcolor{gray}72.32 & 71.09 & 70.77 & 68.44 & 67.25 & 62.61 & 58.77 & 38.76 \\
			160 & 71.51 & 72.15 & 71.61 & 72.12 & \cellcolor{gray}71.04 & \cellcolor{gray}71.31 & 69.36 & 68.60 & 64.69 & 61.53 & 43.02 \\
			128 & 69.04 & 70.30 & 69.78 & 70.90 & 70.19 & 71.01 & \cellcolor{gray}69.35 & \cellcolor{gray}69.28 & 65.96 & 63.87 & 47.33 \\
			96 & 62.96 & 65.52 & 65.37 & 67.65 & 66.89 & 68.89 & 67.60 & 68.73 & \cellcolor{gray}65.78 & \cellcolor{gray}65.30 & \cellcolor{gray}51.32 \\
		\end{tabular}
	}
	\resizebox{.5\linewidth}{!}{
		\begin{tabular}{c|ccccccccccc}
			Train $\backslash$ Test & 256 & 224 & 208 & 192 & 176 & 160 & 144 & 128 & 112 & 96 & 64 \\
			\midrule[0.1em]
			224 & \cellcolor{gray}73.04 & \cellcolor{gray}72.19 & 71.13 & 70.09 & 68.48 & 66.10 & 63.13 & 58.63 & 52.50 & 43.91 & 21.23 \\
			192 & 72.67 & 72.26 & \cellcolor{gray}71.47 & \cellcolor{gray}71.08 & 69.68 & 68.18 & 65.84 & 61.61 & 56.86 & 48.81 & 25.37 \\
			160 & 70.83 & 71.39 & 70.17 & 71.08 & \cellcolor{gray}69.31 & \cellcolor{gray}69.50 & 66.63 & 65.05 & 59.60 & 54.08 & 30.56 \\
			128 & 66.70 & 68.18 & 68.19 & 69.14 & 68.65 & 68.98 & \cellcolor{gray}67.78 & \cellcolor{gray}66.74 & 63.73 & 58.69 & 37.14 \\
			96 & 51.48 & 56.73 & 57.06 & 60.85 & 61.13 & 63.99 & 63.55 & 65.13 & \cellcolor{gray}63.40 & \cellcolor{gray}62.70 & \cellcolor{gray}46.15 \\
		\end{tabular}
	}
	\resizebox{.5\linewidth}{!}{
		\begin{tabular}{c|ccccccccccc}
			Train $\backslash$ Test & 256 & 224 & 208 & 192 & 176 & 160 & 144 & 128 & 112 & 96 & 64 \\
			\midrule[0.1em]
			224 & \cellcolor{gray}73.26 & \cellcolor{gray}72.80 & 72.24 & 71.47 & 70.35 & 68.64 & 66.50 & 63.08 & 58.43 & 51.01 & 28.64 \\
			192 & 72.91 & 72.86 & \cellcolor{gray}72.42 & \cellcolor{gray}72.22 & 71.33 & 70.26 & 68.43 & 65.83 & 61.81 & 55.44 & 33.24 \\
			160 & 71.68 & 72.16 & 71.79 & 72.31 & \cellcolor{gray}71.63 & \cellcolor{gray}71.16 & 69.67 & 67.96 & 64.61 & 59.62 & 38.75 \\
			128 & 68.61 & 70.13 & 69.94 & 71.16 & 70.51 & 70.94 & \cellcolor{gray}69.80 & \cellcolor{gray}69.14 & 66.55 & 63.05 & 44.84 \\
			96 & 60.55 & 64.11 & 64.21 & 66.91 & 66.80 & 68.58 & 67.82 & 68.41 & \cellcolor{gray}66.59 & \cellcolor{gray}65.07 & \cellcolor{gray}50.41 \\
		\end{tabular}
	}
	\resizebox{.5\linewidth}{!}{
		\begin{tabular}{c|ccccccccccc}
			Train $\backslash$ Test & 256 & 224 & 208 & 192 & 176 & 160 & 144 & 128 & 112 & 96 & 64 \\
			\midrule[0.1em]
			224 & \cellcolor{gray}77.95 & \cellcolor{gray}77.15 & 76.43 & 75.47 & 74.29 & 72.44 & 70.32 & 66.85 & 62.38 & 54.84 & 32.10 \\
			192 & 77.56 & 77.26 & \cellcolor{gray}76.81 & \cellcolor{gray}76.41 & 75.27 & 74.39 & 72.32 & 69.77 & 65.70 & 59.22 & 37.23 \\
			160 & 76.69 & 76.93 & 76.60 & 76.61 & \cellcolor{gray}75.81 & \cellcolor{gray}75.31 & 73.91 & 72.12 & 69.11 & 63.87 & 43.74 \\
			128 & 74.14 & 75.13 & 75.16 & 75.66 & 75.46 & 75.53 & \cellcolor{gray}74.63 & \cellcolor{gray}73.53 & 71.58 & 67.67 & 51.40 \\
			96 & 68.81 & 70.97 & 71.29 & 72.68 & 72.66 & 73.64 & 73.43 & 73.53 & \cellcolor{gray}72.35 & \cellcolor{gray}70.61 & \cellcolor{gray}59.09 \\
		\end{tabular}
	}
	\resizebox{.5\linewidth}{!}{
		\begin{tabular}{c|ccccccccccc}
			Train $\backslash$ Test & 256 & 224 & 208 & 192 & 176 & 160 & 144 & 128 & 112 & 96 & 64 \\
			\midrule[0.1em]
			224 & \cellcolor{gray}78.93 & \cellcolor{gray}78.57 & 78.32 & 77.81 & 77.02 & 76.11 & 74.67 & 72.47 & 69.44 & 64.30 & 44.91 \\
			192 & 78.63 & 78.67 & \cellcolor{gray}78.43 & \cellcolor{gray}78.44 & 77.64 & 77.10 & 75.86 & 74.00 & 71.29 & 67.05 & 48.91 \\
			160 & 77.66 & 78.07 & 77.99 & 78.26 & \cellcolor{gray}77.73 & \cellcolor{gray}77.60 & 76.60 & 75.35 & 73.15 & 69.49 & 53.62 \\
			128 & 75.18 & 76.22 & 76.39 & 77.17 & 77.01 & 77.38 & \cellcolor{gray}76.50 & \cellcolor{gray}76.04 & 74.24 & 71.56 & 57.84 \\
			96 & 69.91 & 72.24 & 72.69 & 74.18 & 74.33 & 75.39 & 75.12 & 75.30 & \cellcolor{gray}74.28 & \cellcolor{gray}72.83 & \cellcolor{gray}62.14 \\
		\end{tabular}
	}
	\vskip -0.1in
\end{table}

\begin{table}[htbp]
	\caption{Comparison of ResNet-18-based SAN, evaluated on interpolated resolutions with different inference configurations, including (i) ideal inference: before (\textit{left}) and after (\textit{middle}) BN calibration; (ii) proxy inference: no need for BN calibration (\textit{right}).}
	\label{tab:bn-calibration}
	\vskip -0.1in
	\resizebox{.32\linewidth}{!}{
		\begin{tabular}{c|cccc}
			Train $\backslash$ Test & 208 & 176 & 144 & 112 \\
			\midrule[0.1em]
			224 & 71.71 & 67.44 & 57.47 & 39.46 \\
			192 & \cellcolor{gray}71.51 & 70.72 & 65.22 & 52.15 \\
			160 & 66.21 & \cellcolor{gray}70.52 & 69.07 & 61.10 \\
			128 & 49.45 & 64.49 & \cellcolor{gray}68.78 & 65.57 \\
			96 & 19.71 & 45.75 & 61.74 & \cellcolor{gray}65.26 \\
		\end{tabular}
	}
	\resizebox{.32\linewidth}{!}{
		\begin{tabular}{c|cccc}
			Train $\backslash$ Test & 208 & 176 & 144 & 112 \\
			\midrule[0.1em]
			224 & 72.03 & 70.32 & 66.90 & 59.18 \\
			192 & \cellcolor{gray}72.13 & 71.13 & 68.48 & 62.14 \\
			160 & 71.41 & \cellcolor{gray}71.07 & 69.40 & 64.59 \\
			128 & 68.95 & 69.89 & \cellcolor{gray}69.52 & 66.16 \\
			96 & 62.49 & 65.86 & 67.35 & \cellcolor{gray}66.26 \\
		\end{tabular}
	}
	\resizebox{.32\linewidth}{!}{
		\begin{tabular}{c|cccc}
			Train $\backslash$ Test & 208 & 176 & 144 & 112 \\
			\midrule[0.1em]
			224 & 71.89 & 70.39 & 67.15 & 59.96 \\
			192 & \cellcolor{gray}72.10 & 71.09 & 68.44 & 62.61 \\
			160 & 71.61 & \cellcolor{gray}71.04 & 69.36 & 64.69 \\
			128 & 69.78 & 70.19 & \cellcolor{gray}69.35 & 65.96 \\
			96 & 65.37 & 66.89 & 67.60 & \cellcolor{gray}65.78 \\
		\end{tabular}
	}
	\vskip -0.1in
\end{table}

Several different choices of main networks are explored to demonstrate the effectiveness and scalability of our approach. Specifically, we select three network architectures including ResNet~\cite{He_2016_CVPR} with 18/50 layers and MobileNetV2~\cite{Sandler_2018_CVPR} in view of their strong track record. It is noted that we consider both large-scale networks and a very lightweight one, which also feature regular and inverted residual blocks respectively. Furthermore, we prove that our meta learners could smoothly learn to generate the kernels of both standard convolutions and depthwise separable convolutions. The ResNet family is trained using the default SGD optimizer with the momentum of 0.9 and the weight decay of 1e-4 for 120 epochs. The learning rate initiates from 0.1 and decays to zero following a half cosine annealing schedule. The batch size is set to 256. The lightweight MobileNetV2 is trained using almost the same optimization hyperparameters and learning rate decay strategy but with a smaller initial learning rate of 0.05 and a smaller weight decay of 4e-5 to suppress underfitting. The optimization procedure lasts for 150 epochs for full convergence. All baselines and our SAN-based models are trained using the above scheme for fair comparisons. The training resolutions of SAN models are set to $\mathbb{S}=\{S_1, S_2, S_3, S_4, S_5\}=\{224,192,160,128,96\}$.

We take independently trained models as the baselines and evaluate them among a wide range of test resolutions on the ImageNet validation set. The baseline results of three networks are shown in the left panel of Table~\ref{tab:san-comparison}. We report corresponding results using our SAN in the right panel of Table~\ref{tab:san-comparison}, where the results in the $j$\textsuperscript{th} row is evaluated with the scale encoding $\varepsilon^{(S_j)}$ and BN layers $\text{BN}^{(S_j)}$ without the calibration of statistics. Therefore, the shaded numeric value (in the $j$\textsuperscript{th} row, s.t. $S(T)=S_j$) points out the inference result using the \textit{proxy inference} method for the test resolution $T$ in each column, as stated in Section~\ref{sec:inference}. It is evident that our dynamically parameterized models achieve consistent accuracy improvement over the individually trained baselines. Such performance enhancement emerges not only on the training resolutions but also on the interpolated and extrapolated ones of the training range, demonstrating the universal applicability of our meta learner. Furthermore, it should be noticed that the generated classifier for one specific resolution also generalizes well on other resolutions (indicated by those numeric values without being shaded in each row) compared to the individually trained model, even obtaining over 10\% compensation for those baseline models in several cases with severe scale deviation (around the corner of the table).

The recognition accuracy curves of our SAN with MobileNetV2 and corresponding baseline models are depicted in Fig.~\ref{fig:envelop}. It showcases a clear trend that the curves of SAN models envelop those of their baseline counterparts. Our SAN also guarantees much milder performance drop when input samples pose a resolution discrepancy between optimization and inference.

\subsection{Ablation Studies}
We conduct comprehensive ablation experiments to analyze the influence of different configurations and provide empirical validation for our design.

\textbf{BN Calibration for Ideal Inference.}
The encoding scalars used for training are bound to their privatized BN layers. If these scalars are altered according to input resolutions at test time following the ideal inference method, evaluation performance might suffer from incompatibility, as demonstrated in the left panel of Table~\ref{tab:bn-calibration}. For compensation, we apply the post-hoc calibration for BN statistics. Specifically, we recalculate the running mean and variance of BN layers over training samples of the test resolution with exact averages rather than moving averages. The evaluation performance after BN calibration is reported in the middle panel of Table~\ref{tab:bn-calibration}, showing amelioration for those before calibration. To deserve to be mentioned, when the training resolution approaches the test one, the performance gap would be relatively minor, crediting to the smoothness of the linear meta modeling space and the BN parameter space. Anyway, the above dissection reiterates the critical value of Batch Normalization.

\textbf{Proxy Inference.}
We empirically justify that using $\boldsymbol{\theta}_{\text{proxy}}$ for inference is a credible alternative for model evaluation. For test resolutions included in the training resolution range, \textit{i}.\textit{e}., $224, 192, 160, 128$ and $96$, $S(T)=T$, thus $\boldsymbol{\theta}_{\text{proxy}}=\boldsymbol{\theta}_{\text{ideal}}$ and the inference process will be identical for the two proposed methods. Thus, we lay analytic emphasis on the test resolutions sandwiched between two training resolutions, \textit{i}.\textit{e}., $208, 176, 144$ and $112$, and report their performance in the right panel of Table~\ref{tab:bn-calibration}. Since the scale encoding reverts to the ones during training in the proxy inference method, calibration is not needed here. We notice that using proxy inference for each test resolution (as shaded in the right panel of Table~\ref{tab:bn-calibration}) leads to nearly the same results as those with the ideal inference method (as shaded in the middle panel of Table~\ref{tab:bn-calibration}). These comparisons provide empirical support for our claim in Section~\ref{sec:inference}.

\begin{table}[htbp]
	\caption{Accuracy of MobileNetV2-based SAN (w/o SD) (\textit{left}) and MobileNetV2-based SAN (w/o SD) with shared Batch Normalization layers (\textit{right}). Please refer to Table~\ref{tab:san-comparison} (\textit{middle-right}) for accuracy of intact MobileNetV2-based SAN.}
	\vskip -0.1in
	\centering
	\label{tab:mobilenetv2-meta}
	\resizebox{.49\linewidth}{!}{
		\begin{tabular}{c|ccccccccccc}
			Train $\backslash$ Test & 256 & 224 & 208 & 192 & 176 & 160 & 144 & 128 & 112 & 96 & 64 \\
			\midrule[0.1em]
			224 & 73.14 & 72.53 & 71.50 & 70.94 & 69.26 & 67.91 & 64.50 & 61.84 & 55.18 & 48.60 & 24.94 \\
			192 & 72.85 & 72.58 & 71.45 & 71.56 & 69.93 & 69.36 & 66.32 & 64.15 & 58.23 & 52.52 & 28.62 \\
			160 & 71.30 & 71.78 & 70.66 & 71.54 & 69.92 & 70.00 & 67.35 & 66.09 & 60.85 & 56.20 & 32.94 \\
			128 & 67.42 & 69.11 & 67.64 & 69.95 & 68.37 & 69.65 & 67.43 & 67.19 & 62.64 & 59.54 & 38.26 \\
			96 & 56.25 & 60.34 & 58.50 & 63.81 & 61.86 & 66.25 & 63.87 & 66.23 & 62.25 & 61.86 & 44.05 \\
		\end{tabular}
	}
	\label{tab:mobilenetv2-meta-shared-bn}
	\resizebox{.5\linewidth}{!}{
		\begin{tabular}{c|ccccccccccc}
			Train $\backslash$ Test & 256 & 224 & 208 & 192 & 176 & 160 & 144 & 128 & 112 & 96 & 64 \\
			\midrule[0.1em]
			224 & \hspace{0.5em}2.82 & \hspace{0.5em}2.95 & \hspace{0.5em}2.88 & \hspace{0.5em}3.02 & \hspace{0.5em}2.93 & \hspace{0.5em}3.04 & \hspace{0.5em}2.90 & \hspace{0.5em}2.77 & \hspace{0.5em}2.59 & \hspace{0.5em}2.29 & \hspace{0.5em}1.13 \\
			192 & 27.27 & 29.07 & 29.03 & 29.91 & 29.30 & 29.24 & 28.09 & 25.70 & 23.24 & 18.64 & \hspace{0.5em}8.45 \\
			160 & 66.41 & 66.89 & 66.46 & 66.52 & 65.62 & 65.02 & 63.08 & 60.61 & 56.812 & 50.66 & 29.64 \\
			128 & 49.17 & 50.86 & 50.26 & 51.43 & 50.10 & 50.70 & 47.92 & 46.73 & 41.52 & 37.71 & 19.82 \\
			96 & \hspace{0.5em}0.76 & \hspace{0.5em}0.90 & \hspace{0.5em}0.85 & \hspace{0.5em}1.03 & \hspace{0.5em}0.94 & \hspace{0.5em}1.23 & \hspace{0.5em}1.13 & \hspace{0.5em}1.28 & \hspace{0.5em}1.07 & \hspace{0.5em}1.38 & \hspace{0.5em}1.07 \\
		\end{tabular}
	}
	\vskip -0.1in
\end{table}
\begin{table}[htbp]
	\caption{Accuracy of parallel ResNet-18 models with privatized convolution for each resolution and scale distillation across resolutions. Please refer to Table~\ref{tab:san-comparison} (\textit{top-left} and \textit{top-right}) for results of the baseline and SAN.}
	\vskip -0.1in
	\centering
	\label{tab:resnet18-privatized-conv}
	\resizebox{.49\linewidth}{!}{
		\begin{tabular}{c|ccccccccccc}
			Train $\backslash$ Test & 256 & 224 & 208 & 192 & 176 & 160 & 144 & 128 & 112 & 96 & 64 \\
			\midrule[0.1em]
			224 & 72.06 & 71.10 & 69.54 & 68.87 & 66.23 & 64.54 & 60.47 & 56.89 & 50.11 & 42.60 & 20.32 \\
			192 & 72.53 & 71.94 & 70.41 & 70.79 & 68.28 & 67.93 & 64.17 & 61.97 & 55 33 & 49.39 & 25.96 \\
			160 & 70.93 & 71.37 & 70.34 & 71.21 & 69.55 & 69.82 & 66.94 & 65.62 & 60.57 & 55.92 & 33.51 \\
			128 & 67.57 & 69.11 & 68.56 & 69.89 & 68.78 & 69.89 & 67.81 & 67.87 & 63.84 & 61.46 & 41.96 \\
			96 & 56.80 & 60.81 & 60.75 & 64.31 & 63.53 & 66.67 & 65.22 & 67.32 & 64.35 & 64.53 & 50.62 \\
		\end{tabular}
	}
	\vskip -0.1in
\end{table}

\textbf{Scale Distillation.} For the purpose of ablating the influence of our proposed Scale Distillation, we further train a MobileNetV2-based SAN without this technique. Accuracy records of the original SAN are presented in Table~\ref{tab:san-comparison} (middle-right) and the accuracy of SAN (w/o SD) is provided in Table~\ref{tab:mobilenetv2-meta} for comparison. We observe that the accuracy drop increases in the test scenarios with smaller input scales, which empirically validates the importance of transferring knowledge from high resolutions to lower ones by introducing the extra supervision signal. As a side benefit, the performance of models over interpolated resolutions is ameliorated to a large extent after the application of Scale Distillation, which may be attributed to the improved interaction of multi-scale information among the same image instances with different resolutions.

\textbf{Privatized BN.} Sharing similar philosophy with~\cite{touvron2019FixRes} and~\cite{yu2018slimmable}, we privatize BN layers by design to eliminate incompatibility between various scaled distributions. For the purpose of ablation, we take the above MobileNetV2-based SAN (w/o SD) as an exemplar and substitute shared BN layers for privatized ones. The consequent accuracy is provided in Table~\ref{tab:mobilenetv2-meta-shared-bn}, where the benefit of BN is skewed to resolutions in the middle of the training resolution range while the performance in each case still deteriorates compared to its counterpart with privatized BNs. We observe that the training process is stable but the validation accuracy is extremely low because during training the mean and variance of the current mini-batch with a specific resolution are applied while the validation process depends on the moving average statistics from all resolutions. The experimental results further prove the point that a unified BN is insufficient to strike a balance among a broad range of resolutions concurrently.

\textbf{Parameterized Convolution.} A quick question is that why not simply use privatized convolutional layers for each main network rather than employing an auxiliary meta learner? We take the ResNet18 model as an example and show the accuracy with privatized convolutional layers in Table~\ref{tab:resnet18-privatized-conv} (scale distillation is still applicable in this context). In comparison, the performance are mostly superior to baseline but inferior to our proposed SAN. The meta learning algorithm is adept at integrating prior experience from multiple existing tasks and achieving fast adaption to unseen tasks. In accordance to this rationale, the comparison results also speak to the advantage of utilizing meta networks to collect multi-scale information and dynamically parameterize the convolutional layers for arbitrary image resolutions at runtime.

\begin{table}[htbp]
	\caption{Absolute accuracy drop of ResNet-18-based SAN with non-linear meta learners using 8/16 (\textit{left}/\textit{right}) hidden units in the FC layer. Please refer to Table~\ref{tab:san-comparison} (\textit{top-right}) for results of the original SAN equipped with linear meta learners. Note that negative numbers indicate accuracy increment.}
	\vskip -0.1in
	\label{tab:resnet18-meta-distill-non-linear-8}
	\centering
	\resizebox{.49\linewidth}{!}{
		\begin{tabular}{c|ccccccccccc}
			Train $\backslash$ Test & 256 & 224 & 208 & 192 & 176 & 160 & 144 & 128 & 112 & 96 & 64 \\
			\midrule[0.1em]
			224 & -0.22 & -0.33 & 0.11 & 0.12 & 0.15 & 0.01 & 0.26 & 0.19 & 0.29 & 0.76 & 1.50 \\
			192 & -0.21 & -0.07 & 0.19 & 0.25 & 0.27 & 0.20 & 0.36 & 0.56 & 0.92 & 1.30 & 2.27 \\
			160 & -0.08 & -0.03 & 0.18 & 0.02 & 0.19 & 0.14 & 0.27 & 0.29 & 0.62 & 0.88 & 2.16 \\
			128 & 0.22 & 0.19 & 0.16 & -0.02 & 0.43 & 0.09 & 0.22 & -0.00 & 0.25 & 0.46 & 1.24 \\
			96 & 0.93 & 0.80 & 1.00 & 0.54 & 0.76 & 0.36 & 0.66 & 0.26 & 0.09 & 0.11 & -0.07 \\
		\end{tabular}
	}
	\label{tab:resnet18-meta-distill-non-linear-16}
	\centering
	\resizebox{.49\linewidth}{!}{
		\begin{tabular}{c|ccccccccccc}
			Train $\backslash$ Test & 256 & 224 & 208 & 192 & 176 & 160 & 144 & 128 & 112 & 96 & 64 \\
			\midrule[0.1em]
			224 & -0.48 & -0.31 & -0.24 & 0.01 & -0.17 & -0.11 & 0.21 & 0.67 & 0.55 & 1.35 & 2.20 \\
			192 & -0.39 & -0.13 & -0.03 & 0.09 & -0.13 & 0.08 & 0.34 & 1.03 & 1.04 & 1.80 & 2.74 \\
			160 & -0.33 & -0.18 & -0.05 & -0.11 & -0.23 & 0.00 & 0.02 & 0.42 & 0.57 & 1.25 & 2.27 \\ 
			128 & 0.11 & 0.11 & -0.11 & -0.06 & -0.12 & 0.08 & -0.26 & 0.03 & 0.03 & 0.42 & 1.32 \\
			96 & 1.26 & 1.04 & 0.87 & 0.74 & 0.27 & 0.38 & 0.02 & 0.23 & -0.28 & 0.12 & 0.05 \\
		\end{tabular}
	}
	\vskip -0.25in
\end{table}

\textbf{Architecture of Meta Learner.} We explore different MLP structures of the meta network design. Besides the single FC layer adopted in our main experiments, we draw into the non-linear activation to construct a two-layer MLP, where consecutive linear FC layers are interleaved with a non-linear ReLU activation operation and the number of hidden units is set to 8 or 16 in order to avoid introducing heavy computational overheads. The ResNet-18 model is selected for this experiment in view of its relatively low memory consumption. The absolute accuracy drop after substituting the non-linear meta leaner with 8 hidden units for the default linear one is summarized in Table~\ref{tab:resnet18-meta-distill-non-linear-8} (left). We could find that the final performance of these two designs is in close proximity to each other on the medium resolutions, which implies that though conceptually simple, a linear modeling space of meta networks is sufficient for giving rise to desired scale-adaptive characteristics and satisfactory performance. Nevertheless, the decline is not negligible on quite large and small resolutions, which might be attributed to over-fitting problems caused by expansion of the non-linear modeling space. It is probably difficult for the SAN (w/ non-linearity) to generalize well to these situations with large scale deviation. We also explore larger hidden-unit space via doubling its unit number and observe that the tendency of overfitting to marginal resolutions becomes more obvious, as shown in Table~\ref{tab:resnet18-meta-distill-non-linear-16} (right).

\textbf{Generalization to Large Resolutions.} We shift the range of training resolutions to a larger magnitude, using ResNet-18 as the test case. As shown in Table~\ref{tab:resnet18-large}, the accuracy improvement of SAN presents a similar trend within a range of large resolutions $\mathbb{S}=\{384,320,256,224,192\}$ compared to the smaller ones discussed in Section~\ref{sec:main}. The results empirically show the effectiveness and robustness of our method in handling images with different ranges of resolutions.

\textbf{Comparison to FixRes.} FixRes~\cite{touvron2019FixRes} is merely akin to us in the motivation of alleviating \textit{scale deviation}, but it focuses on further pushing the performance upper boundary of large-scale models with resolution adaption. In stark contrast, we pay attention to enabling efficient and flexible inference conditioned on the input image resolution for evaluation at runtime. A performance comparison is applicable for these two methods using the ResNet-50 model trained on the image resolution of $224 \times 224$, as summarized in Table~\ref{tab:fixres}. We are able to beat FixRes when test resolutions are relatively small, which is especially beneficial for resource-limited environments. Furthermore, due to the limitation of available computational resources now, our training resolution range could be at most set to $\{224,192,160,128,96\}$. If we could enlarge this range, \textit{e}.\textit{g}., including training images with larger resolutions such as 384, 448 and 480, our SAN may achieve better performance on these large scales.

\begin{table}[htbp]
	\caption{Comparison of ResNet-18 baseline models (\textit{top}) and ResNet-18-based SAN with a range of larger training resolutions (\textit{bottom}). The shaded numeric values have the same meaning as Table~\ref{tab:san-comparison}.}
	\vskip -0.1in
	\label{tab:resnet18-baseline-large}
	\centering
	\resizebox{.7\linewidth}{!}{
		\begin{tabular}{c|ccccccccccccc}
			Train $\backslash$ Test & 384 & 320 & 256 & 224 & 208 & 192 & 176 & 160 & 144 & 128 & 112 & 96 & 64 \\
			\midrule[0.1em]
			384 & \cellcolor{gray}73.20 & 71.23 & 67.33 & 63.83 & 60.65 & 58.62 & 54.71 & 50.58 & 45.09 & 38.86 & 32.05 & 23.96 & 8.89 \\
			320 & 73.42 & \cellcolor{gray}72.56 & 70.03 & 67.44 & 64.92 & 63.23 & 59.47 & 56.07 & 51.17 & 45.89 & 38.64 & 30.27 & 11.74 \\
			256 & 72.75 & 72.68 & \cellcolor{gray}71.71 & 70.24 & 68.00 & 67.27 & 64.05 & 62.12 & 57.20 & 53.27 & 45.45 & 38.00 & 17.00 \\
			224 & 71.87 & 72.36 & 71.90 & \cellcolor{gray}70.97 & 69.14 & 68.72 & 66.20 & 64.68 & 60.36 & 56.85 & 50.17 & 42.55 & 20.40 \\
			192 & 69.95 & 71.07 & 71.71 & 71.25 & \cellcolor{gray}69.64 & \cellcolor{gray}69.75 & \cellcolor{gray}67.69 & \cellcolor{gray}67.04 & \cellcolor{gray}63.23 & \cellcolor{gray}60.47 & \cellcolor{gray}54.21 & \cellcolor{gray}47.73 & \cellcolor{gray}25.10 \\
		\end{tabular}
	}
	\label{tab:resnet18-large}
	\centering
	\resizebox{.7\linewidth}{!}{
		\begin{tabular}{c|ccccccccccccc}
			Train $\backslash$ Test & 384 & 320 & 256 & 224 & 208 & 192 & 176 & 160 & 144 & 128 & 112 & 96 & 64 \\
			\midrule[0.1em]
			384 & \cellcolor{gray}74.69 & 74.12 & 72.14 & 70.28 & 68.65 & 67.06 & 64.77 & 61.97 & 58.25 & 53.41 & 47.33 & 39.51 & 18.66 \\
			320 & 74.61 & \cellcolor{gray}74.50 & 73.28 & 71.72 & 70.39 & 69.07 & 67.10 & 64.64 & 62.28 & 56.94 & 51.22 & 43.63 & 21.85 \\
			256 & 73.41 & 74.05 & \cellcolor{gray}73.71 & 72.87 & 71.85 & 70.86 & 69.43 & 67.53 & 64.42 & 60.86 & 55.56 & 48.26 & 25.80 \\
			224 & 72.27 & 73.39 & 73.59 & \cellcolor{gray}73.00 & 72.29 & 71.52 & 70.16 & 68.52 & 65.64 & 62.32 & 57.36 & 50.54 & 28.13 \\
			192 & 70.44 & 72.15 & 73.11 & 72.73 & \cellcolor{gray}72.21 & \cellcolor{gray}71.74 & \cellcolor{gray}70.72 & \cellcolor{gray}69.36 & \cellcolor{gray}66.74 & \cellcolor{gray}63.80 & \cellcolor{gray}59.20 & \cellcolor{gray}52.72 & \cellcolor{gray}30.52 \\
		\end{tabular}
	}
\end{table}
\begin{table}[htbp]
	\caption{Comparison of FixRes and our SAN using ResNet-50 with training image size of $224 \times 224$ on ImageNet. The results of FixRes are extracted from the original publication. The better result for each test resolution is highlighted in \textbf{bold}.}
	\label{tab:fixres}
	\vskip -0.1in
	\centering
	\resizebox{.7\linewidth}{!}{
		\begin{tabular}{c|ccccccc}
			Method $\backslash$ Test resolution & 64 & 128 & 224 & 288 & 384 & 448 & 480 \\
			\midrule[0.1em]
			FixRes~\cite{touvron2019FixRes} & 41.7 & 67.7 & 77.1 & \textbf{78.6} & \textbf{79.0} & \textbf{78.4} & \textbf{78.1} \\
			\textbf{SAN (ours)} & \textbf{44.9} & \textbf{72.5} & \textbf{78.4} & \textbf{78.6} & 77.0 & 75.3 & 74.3 \\
		\end{tabular}
	}
	\vskip -0.1in
\end{table}

\textbf{More Ablation and Analysis.} We provide a great many additional results in the supplementary materials, including analysis of complementary information across resolutions, the superiority of switching input resolutions over network widths, border and round-off effects, results of inference on more dense sampled resolutions, visualization of privatized BN layers, a data-free form of ideal inference, comparison to stronger baselines, and so on.

\section{Conclusion}

In this paper, we have proposed a Scale Adaptive Network (SAN) framework. For each main network encompassed into the meta learning algorithm, convolutional kernels are dynamically generated by meta networks and BN layers are specially treated. The meta learner could parameterize neural networks for visual classification conditioned on the input resolution at runtime, achieving considerably better performance compared with individually trained models. It is compatible with any CNN backbone architectures, providing an adaptive resolution-accuracy trade-off for fast adaption to environments with different real-time demands. The generality of the proposed framework makes it promising to be translated well to other application domains, such as object detection and semantic segmentation, which is expected to appear in the future work.

\section*{Acknowledgement}

We sincerely thank Ming Lu and Aojun Zhou from Intel Labs China for their insightful discussions, warm encouragement and valuable feedback. We would like to thank SenseTime Group Limited for their support of computational resource.

\section*{\LARGE Appendix}
\appendix

\section{Algorithmic Description}

We summarize the optimization pipeline of our SAN framework in Algorithm~\ref{alg:opt}. The two methods for inference mentioned in the Section 3.3 of the main paper are also presented in Algorithm~\ref{alg:ideal} and~\ref{alg:proxy} respectively.

\noindent
{\small
	\IncMargin{1em}
	\begin{algorithm}[H]
		
		\SetAlgoNoLine
		
		\KwIn{
			Training dataset $\mathcal{D}$, maximal iteration $M$, training resolution range $\mathbb{S}=\{S_1, S_2, \cdots, S_k\}$, main network architecture $\mathcal{N}$.\\}
		\KwOut{
			Parameters of the meta network $\mathcal{M}$, $\text{BN}^{(S_i)}$ for each input resolution $S_i, i=1, 2, \cdots, k$ and the FC layer.}
		\BlankLine
		
		Initialize the network parameters $\mathcal{M}$, $\text{BN}^{(S_i)}, i=1, 2, \cdots, k$ and FC. \\
		\For{$t \leftarrow \text{1 to M}$}{
			Randomly draw a batch of samples $\mathcal{B}$ from $\mathcal{D}$. \\
			\ForEach{$\mathbf{X}$ in $\mathcal{B}$}{
				\For{$i \leftarrow \text{1 to k}$}{
					Randomly crop and resize the mini-batch data to the training resolution $S_i$, obtaining $\mathcal{B}^{(S_i)}$ with the prereserved labels. \\
					Derive scale encoding $\varepsilon^{(S_i)}$. \\
					Generate convolutional layers $\mathcal{M}(\varepsilon^{(S_i)})$  via Eqn. 2\footnote{All the referred equations in this algorithm are presented in the main paper.}. \\
					Parameterize $\mathcal{N}^{(S_i)}$ with $\mathcal{M}(\varepsilon^{(S_i)})$, $\text{BN}^{(S_i)}$ and FC. \\
					Foward and compute cross-entropy loss $\mathcal{L}_{\text{CE}}^{(S_i)}$ via Eqn. 1. \\
					Compute scale distillation loss $\mathcal{L}_{\text{SD}}^{(S_i)}$ via Eqn. 3.
				}
				Compute the total loss via an un-weighted summation in Eqn. 4. \\
				Update network parameters with the SGD optimizer.
			}
		}
		\Return $\mathcal{M}$, $\text{BN}^{(S_i)}, i=1, 2, \cdots, k$ and FC.
		\caption{Optimization of Scale Adaptive Network}
		\label{alg:opt}
	\end{algorithm}
	\DecMargin{1em}
}
\clearpage

\noindent
{\small
	\IncMargin{1em}
	\begin{algorithm}[H]
		
		\SetAlgoNoLine
		
		\KwIn{
			Test image $\mathcal{I}$ with the resolution $T$, training dataset $\mathcal{D}$, training resolution range $\mathbb{S}=\{S_1, S_2, \cdots, S_k\}$, main network architecture $\mathcal{N}$, meta network $\mathcal{M}$, $\text{BN}^{(S_i)}$ for each training resolution $S_i, i=1, 2, \cdots, k$ and FC.\\}
		\KwOut{
			Category prediction of the test image.}
		\BlankLine
		
		Search the nearest resolution $S(T) \in \mathbb{S}$ for $T$. \\
		Derive scale encoding $\varepsilon^{(T)}$. \\
		Generate convolutional layers $\mathcal{M}(\varepsilon^{(T)})$ for main network $\mathcal{N}^{(T)}$. \\
		Calibrate statistics of $\text{BN}^{(S(T))}$ using data set $\mathcal{D}^{(T)}$ with test resolution $T$. \\
		Parameterize $\mathcal{N}^{(T)}$ with $\mathcal{M}(\varepsilon^{(T)})$, calibrated $\text{BN}^{(S(T))}$ and FC. \\
		Feed the image $\mathcal{I}$ to $\mathcal{N}^{(T)}$ to get its category prediction $\mathcal{N}^{(T)}(\mathcal{I})$. \\
		\Return $\mathcal{N}^{(T)}(\mathcal{I})$.
		\caption{Ideal Inference of Scale Adaptive Network}
		\label{alg:ideal}
	\end{algorithm}
	\DecMargin{1em}
}

\noindent
{\small
	\IncMargin{1em}
	\begin{algorithm}[H]
		
		\SetAlgoNoLine
		
		\KwIn{
			Test image $\mathcal{I}$ with the resolution $T$, training resolution range $\mathbb{S}=\{S_1, S_2, \cdots, S_k\}$, main network architecture $\mathcal{N}$, meta network $\mathcal{M}$, $\text{BN}^{(S_i)}$ for each training resolution $S_i, i=1, 2, \cdots, k$ and FC.\\}
		\KwOut{
			Category prediction of the test image.}
		\BlankLine
		
		Search the nearest resolution $S(T) \in \mathbb{S}$ for $T$. \\
		Derive scale encoding $\varepsilon^{(S(T))}$\footnote{The steps 2 and 3 can be omitted if the pre-trained model $\mathcal{N}^{(S(T))}$ is on hand.\label{omit}}. \\
		Generate convolutional layers $\mathcal{M}(\varepsilon^{(S(T))})$ for main network $\mathcal{N}^{(S(T))}$\textsuperscript{\ref{omit}}. \\
		Parameterize $\mathcal{N}^{(T)}=\mathcal{N}^{(S(T))}$ with $\mathcal{M}(\varepsilon^{(S(T))})$, $\text{BN}^{(S(T))}$ and FC. \\
		Feed the image $\mathcal{I}$ to $\mathcal{N}^{(T)}$ to get its category prediction $\mathcal{N}^{(T)}(\mathcal{I})$. \\
		\Return $\mathcal{N}^{(T)}(\mathcal{I})$.
		\caption{Proxy Inference of Scale Adaptive Network}
		\label{alg:proxy}
	\end{algorithm}
	\DecMargin{1em}
}

\noindent
{\small
	\IncMargin{1em}
	\begin{algorithm}[H]
		
		\SetAlgoNoLine
		
		\KwIn{
			Test image $\mathcal{I}$ with the resolution $T$, training resolution range $\mathbb{S}=\{S_1, S_2, \cdots, S_k\}$, main network architecture $\mathcal{N}$, meta network $\mathcal{M}$, $\text{BN}^{(S_i)}$ for each training resolution $S_i, i=1, 2, \cdots, k$ and FC.\\}
		\KwOut{
			Category prediction of the test image.}
		\BlankLine
		
		Search the two nearest resolutions $S_{floor}(T), S_{ceil}(T) \in \mathbb{S}$ for $T$, s.t. $S_{floor}(T)< T < S_{ceil}(T)$. \\
		Derive scale encoding $\varepsilon^{(T)}$. \\
		Generate convolutional layers $\mathcal{M}(\varepsilon^{(T)})$ for main network $\mathcal{N}^{(T)}$. \\
		Calculate $\text{BN}^{(T)}$ by linearly interpolating between $\text{BN}^{(S_{floor}(T))}$ and $\text{BN}^{(S_{ceil}(T))}$. \\
		Parameterize $\mathcal{N}^{(T)}$ with $\mathcal{M}(\varepsilon^{(T)})$, $\text{BN}^{(T)}$ and FC. \\
		Feed the image $\mathcal{I}$ to $\mathcal{N}^{(T)}$ to get its category prediction $\mathcal{N}^{(T)}(\mathcal{I})$. \\
		\Return $\mathcal{N}^{(T)}(\mathcal{I})$.
		\caption{Data-Free Ideal Inference of Scale Adaptive Network}
		\label{alg:DF-ideal}
	\end{algorithm}
	\DecMargin{1em}
}

\begin{table}[htbp]
	\caption{Ratios of weights to biases in the meta networks for each individual layer in the ResNet-18 (\textit{left}), ResNet-50 (\textit{middle}) and MobileNetV2 (\textit{right}) models.}
	\label{tab:ratio}
	\centering
	\resizebox{.3\linewidth}{!}{
		\begin{tabular}{c|c|c}
			\toprule[0.1em]
			Block & Layer & Ratio \\
			\midrule[0.1em]
			\multirow{2}*{conv2\_0} & layer1 & 0.4750 \\
			& layer2 & 0.4894 \\
			\hline
			\multirow{2}*{conv2\_1} & layer1 & 0.4821 \\
			& layer2 & 0.4963 \\
			\hline
			\multirow{3}*{conv3\_0} & layer1 & 0.4853 \\
			& layer2 & 0.5026 \\
			& skip connection & 0.4993 \\
			\hline
			\multirow{2}*{conv3\_1} & layer1 & 0.5002 \\
			& layer2 & 0.5081 \\
			\hline
			\multirow{3}*{conv4\_0} & layer1 & 0.4937 \\
			& layer2 & 0.5137 \\
			& skip connection & 0.4766 \\
			\hline
			\multirow{2}*{conv4\_1} & layer1 & 0.5114 \\
			& layer2 & 0.5179 \\
			\hline
			\multirow{3}*{conv5\_0} & layer1 & 0.5012 \\
			& layer2 & 0.5222 \\
			& skip connection & 0.4953 \\
			\hline
			\multirow{2}*{conv5\_1} & layer1 & 0.5263 \\
			& layer2 & 0.5316 \\
			\bottomrule[0.1em]
		\end{tabular}
	}
	\hspace{1.0em}
	\resizebox{.3\linewidth}{!}{
		\begin{tabular}{c|c|c}
			\toprule[0.1em]
			Block & Layer & Ratio \\
			\midrule[0.1em]
			\multirow{4}*{conv2\_0} & layer1 & 0.5117 \\
			& layer2 & 0.4500 \\
			& layer3 & 0.4386 \\
			& skip connection & 0.4652 \\
			\hline
			\multirow{3}*{conv2\_1} & layer1 & 0.4330 \\
			& layer2 & 0.4400 \\
			& layer3 & 0.4521 \\
			\hline
			\multirow{3}*{conv2\_2} & layer1 & 0.4485 \\
			& layer2 & 0.4418 \\
			& layer3 & 0.4501 \\
			\hline
			\multirow{4}*{conv3\_0} & layer1 & 0.4414 \\
			& layer2 & 0.4340 \\
			& layer3 & 0.4367 \\
			& skip connection & 0.4414 \\
			\hline
			\multirow{3}*{conv3\_1} & layer1 & 0.4525 \\
			& layer2 & 0.4600 \\
			& layer3 & 0.4468 \\
			\hline
			\multirow{3}*{conv3\_2} & layer1 & 0.4557 \\
			& layer2 & 0.4574 \\
			& layer3 & 0.4549 \\
			\hline
			\multirow{3}*{conv3\_3} & layer1 & 0.4497 \\
			& layer2 & 0.4602 \\
			& layer3 & 0.4557 \\
			\hline
			\multirow{4}*{conv4\_0} & layer1 & 0.4549 \\
			& layer2 & 0.4401 \\
			& layer3 & 0.4490 \\
			& skip connection & 0.4424 \\
			\hline
			\multirow{3}*{conv4\_1} & layer1 & 0.4731 \\
			& layer2 & 0.4797 \\
			& layer3 & 0.4754 \\
			\hline
			\multirow{3}*{conv4\_2} & layer1 & 0.4676 \\
			& layer2 & 0.4797 \\
			& layer3 & 0.4746 \\
			\hline
			\multirow{3}*{conv4\_3} & layer1 & 0.4705 \\
			& layer2 & 0.4775 \\
			& layer3 & 0.4753 \\
			\hline
			\multirow{3}*{conv4\_4} & layer1 & 0.4699 \\
			& layer2 & 0.4756 \\
			& layer3 & 0.4766 \\
			\hline
			\multirow{3}*{conv4\_5} & layer1 & 0.4705 \\
			& layer2 & 0.4706 \\
			& layer3 & 0.4731 \\
			\hline
			\multirow{4}*{conv5\_0} & layer1 & 0.4691 \\
			& layer2 & 0.4576 \\
			& layer3 & 0.4635 \\
			& skip connection & 0.4609 \\
			\hline
			\multirow{3}*{conv5\_1} & layer1 & 0.4844 \\
			& layer2 & 0.4919 \\
			& layer3 & 0.4940 \\
			\hline
			\multirow{3}*{conv5\_2} & layer1 & 0.4930 \\
			& layer2 & 0.5079 \\
			& layer3 & 0.5108 \\
			\bottomrule[0.1em]
		\end{tabular}
	}
	\hspace{1.0em}
	\resizebox{.25\linewidth}{!}{
		\begin{tabular}{c|c|c}
			\toprule[0.1em]
			Block & \hspace{.5em}Layer\hspace{.5em} & \hspace{.5em}Ratio\hspace{.5em} \\
			\midrule[0.1em]
			block0 & layer1 & 0.4561 \\
			\hline
			\multirow{2}*{block1} & layer1 & 0.5730 \\
			& layer2 & 0.4905 \\
			\hline
			\multirow{3}*{block2} & layer1 & 0.4874 \\
			& layer2 & 0.4946 \\
			& layer3 & 0.4703 \\
			\hline
			\multirow{3}*{block3} & layer1 & 0.5081 \\
			& layer2 & 0.5487 \\
			& layer3 & 0.5362 \\
			\hline
			\multirow{3}*{block4} & layer1 & 0.4949 \\
			& layer2 & 0.4617 \\
			& layer3 & 0.4730 \\
			\hline
			\multirow{3}*{block5} & layer1 & 0.4931 \\
			& layer2 & 0.5195 \\
			& layer3 & 0.5067 \\
			\hline
			\multirow{3}*{block6} & layer1 & 0.5051 \\
			& layer2 & 0.5229 \\
			& layer3 & 0.5054 \\
			\hline
			\multirow{3}*{block7} & layer1 & 0.4720 \\
			& layer2 & 0.5268 \\
			& layer3 & 0.4647 \\
			\hline
			\multirow{3}*{block8} & layer1 & 0.5064 \\
			& layer2 & 0.6210 \\
			& layer3 & 0.5130 \\
			\hline
			\multirow{3}*{block9} & layer1 & 0.5080 \\
			& layer2 & 0.6138 \\
			& layer3 & 0.5095 \\
			\hline
			\multirow{3}*{block10} & layer1 & 0.5103 \\
			& layer2 & 0.6087 \\
			& layer3 & 0.5059 \\
			\hline
			\multirow{3}*{block11} & layer1 & 0.4898 \\
			& layer2 & 0.8277 \\
			& layer3 & 0.4806 \\
			\hline
			\multirow{3}*{block12} & layer1 & 0.5007 \\
			& layer2 & 0.6022 \\
			& layer3 & 0.5089 \\
			\hline
			\multirow{3}*{block13} & layer1 & 0.4992 \\
			& layer2 & 0.6108 \\
			& layer3 & 0.5006 \\
			\hline
			\multirow{3}*{block14} & layer1 & 0.4941 \\
			& layer2 & 0.6202 \\
			& layer3 & 0.4818 \\
			\hline
			\multirow{3}*{block15} & layer1 & 0.5345 \\
			& layer2 & 0.8139 \\
			& layer3 & 0.5334 \\
			\hline
			\multirow{3}*{block16} & layer1 & 0.5284 \\
			& layer2 & 0.7795 \\
			& layer3 & 0.5212 \\
			\hline
			\multirow{3}*{block17} & layer1 & 0.5011 \\
			& layer2 & 1.6666 \\
			& layer3 & 0.4897 \\
			\bottomrule[0.1em]
		\end{tabular}
	}
\end{table}

\section{Ratios of Weights to Biases in the Meta Networks}

As a sanity check, we investigate the trained meta network for the $l$\textsuperscript{th} convolutional layer and evaluate the ratio of its weight $\textbf{W}_l$ to bias $\textbf{b}_l$. Specifically, we take the average of their absolute values and compute the corresponding ratio. We report these derived ratios for each layer of three backbone network architectures in Table~\ref{tab:ratio}. We identify that these ratios are mostly around 0.5, showing a similar level of magnitude between weights and biases, thus the equivalent importance in contributing to the learning dynamics of the meta networks.

\section{Complementary Information across Input Resolutions}

Through the statistical experiments in this section, we demonstrate why diverse input resolutions could contain complementary information and whether our proposed SAN framework takes advantage of such information or not. Regarding the first question, we maintain a suspicion of whether the correctly classified images at a small resolution will fall into the subset of those correctly classified at another large resolution. We conduct relevant experiments using ResNet-18 on ImageNet with the selected training resolution range $\mathbb{S} = \{S_1, S_2, S_3, S_4, S_5\} = \{224, 192, 160, 128, 96\}$ and summarize the statistics in the left panel of Table~\ref{tab:miss-hit}. The numeric value located in the $i$\textsuperscript{th} row and $j$\textsuperscript{th} column refers to the proportion of validation images missed in the top-1 prediction of $\mathcal{N}^{(S_i)}$ but hit in that of $\mathcal{N}^{(S_j)}$. It is consistent with our intuition that the hit rate reduces with the decreasing resolution inside each row. However, it is intriguing that there remain around 5\% hit rates in the upper triangular part of this matrix chart, which implies a negative response to our suspicion above. The inspiration derived from the results is that images resized to one resolution could contain information complementary to their versions of other resolutions regarding the recognition process. Hence it would be promising to enhance the model performance on one test resolution through learning informative features from other resolutions during the mixed-scale training process. Hopefully, the model optimized under this collaborative regime could also improve its tolerance to other auxiliary resolutions in turn during evaluation. In order to justify these conjectures, we also show the improved results of ResNet18-based SAN in the right panel of Table~\ref{tab:miss-hit}. Since the mixed-scale training enforces one image instance to be correctly recognized at a spectrum of resolutions simultaneously, the model predictions achieve better consistency (reduced hit-miss ratios) across different resolutions under our SAN framework, validated by the comparison between the two sub-tables in Table~\ref{tab:miss-hit}. Therefore, the scale-adaptive behavior and improved performance of SAN could be attributed to better utilization of the complementary information across different input resolutions.

\begin{table}[htbp]
	\caption{Percentage of ImageNet validation images which are mistakenly classified at one input resolution but accurately recognized at another resolution according to the top-1 prediction. The ResNet-18-based SAN (\textit{right}) achieves consistently reduced hit-miss ratios compared to the ResNet-18 baseline models (\textit{left}).}
	\label{tab:miss-hit}
	\centering
	\resizebox{.49\linewidth}{!}{
		\begin{tabular}{c|ccccc}
			\toprule[0.2em]
			Miss $\backslash$ Hit & $224 \times 224$ & $192 \times 192$ & $160 \times 160$ & $128 \times 128$ & $96 \times 96$ \\
			\midrule[0.2em]
			$224 \times 224$ & 0 & 5.60\% & 5.50\% & 5.17\% & 4.73\% \\
			$192 \times 192$ & 6.82\% & 0 & 5.88\% & 5.47\% & 5.06\% \\
			$160 \times 160$ & 8.00\% & 7.15\% & 0 & 5.92\% & 5.27\% \\
			$128 \times 128$ & 9.79\% & 8.86\% & 8.04\% & 0 & 5.81\% \\
			$96 \times 96$ & 13.15\% & 12.26\% & 11.19\% & 9.61\% & 0 \\
			\bottomrule[0.2em]
		\end{tabular}
	}
	\resizebox{.49\linewidth}{!}{
		\begin{tabular}{c|ccccc}
			\toprule[0.2em]
			Miss $\backslash$ Hit & $224 \times 224$ & $192 \times 192$ & $160 \times 160$ & $128 \times 128$ & $96 \times 96$ \\
			\midrule[0.2em]
			$224 \times 224$ & 0 & 2.88\% & 3.33\% & 3.67\% & 3.76\% \\
			$192 \times 192$ & 3.81\% & 0 & 3.18\% & 3.57\% & 3.79\% \\
			$160 \times 160$ & 5.88\% & 4.79\% & 0 & 3.45\% & 3.75\% \\
			$128 \times 128$ & 9.02\% & 7.99\% & 6.25\% & 0 & 3.98\% \\
			$96 \times 96$ & 12.38\% & 11.74\% & 10.36\% & 8.54\% & 0 \\
			\bottomrule[0.2em]
		\end{tabular}
	}
\end{table}

\begin{table}[htbp]
	\caption{Top-1 accuracy (\%) of MobileNetV2 baseline models with different width multipliers, S-MobileNetV2 and US-MobileNetV2 (\textit{left}), as well as baseline models operating on different input resolutions and MobileNetV2-based-SAN (\textit{right}) on the ImageNet validation set. The results of S-Net and US-Net are extracted from the official publications.}
	\label{tab:width-resolution}
	\centering
	\resizebox{.525\linewidth}{!}{
		\begin{tabular}{c|c|c|c|c}
			\toprule[0.2em]
			Width Multiplier & MFLOPs & MobileNetV2 & S-Net~\cite{yu2018slimmable} & US-Net~\cite{Yu_2019_ICCV} \\
			\midrule[0.2em]
			$1.0\times$ & 301 & 72.2 & 70.5 & 71.5 \\
			$0.8\times$ & 222 & 69.8 & - & 70.0 \\
			$0.65\times$ & 161 & 68.0 & - & 68.3 \\
			$0.5\times$ & 97 & 64.6 & 64.4 & 65.0 \\
			$0.35\times$ & 59 & 60.1 & 59.7 & 62.2 \\
			\bottomrule[0.2em]
		\end{tabular}
	}
	\resizebox{.465\linewidth}{!}{
		\begin{tabular}{c|c|c|c}
			\toprule[0.2em]
			Input Resolution & MFLOPs & MobileNetV2 & SAN (\textbf{ours}) \\
			\midrule[0.2em]
			$224 \times 224$ & 300 & 72.2 & \textbf{72.8} \\
			$192 \times 192$ & 221 & 71.1 & \textbf{72.2} \\
			$160 \times 160$ & 154 & 69.5 & \textbf{71.2} \\
			$128 \times 128$ & 99 & 66.7 & \textbf{69.1} \\
			$96 \times 96$ & 56 & 62.7 & \textbf{65.1} \\
			\bottomrule[0.2em]
		\end{tabular}
	}
\end{table}

\section{Switching Input Resolution vs. Width Multipliers}

Tuning the width multipliers and input resolutions are two ubiquitous ways to adapt CNNs to the on-device computational budget~\cite{2017arXiv170404861H}\cite{Sandler_2018_CVPR}\cite{Howard_2019_ICCV}. We compare the effect on performance via scaling the width multiplier or input resolution to achieve a similar level of computational complexities, taking MobileNetV2 as the backbone network. From the comparison between baseline MobileNetV2 models in the two sub-tables of Table~\ref{tab:width-resolution}, we reach the preliminary conclusion that adjusting input resolutions is more effective at achieving higher accuracy with the similar computational cost. Recently, a series of methods, S-MobileNetV2~\cite{yu2018slimmable} and US-MobileNetV2~\cite{Yu_2019_ICCV}, were proposed to flexibly switch the network width to achieve an accuracy-efficiency trade-off at runtime, but they merely obtain marginal gains or even slight drop regarding the recognition accuracy as shown in the left panel of Table~\ref{tab:width-resolution}. In contrast, our SAN realizes dynamic and adaptive inference from the perspective of input resolutions, achieving further performance enhancement even compared to those strong baseline models operating on their training resolutions as shown in the right panel of Table~\ref{tab:width-resolution}.

\section{Generalization to Other Large Resolutions}

Similar to an ablation study in the Section 4.2 of the main paper, we also shift the range of training resolutions not so aggressively, with a selection of $\mathbb{S}=\{320,256,224,192,160\}$. The accuracy of baseline models trained individually are summarized in the top panel of Table~\ref{tab:resnet18-baseline-medium} while the accuracy of SAN are shown in the bottom panel of Table~\ref{tab:resnet18-medium}. The tendency of accuracy gains with respect to test resolutions is similar to the results based on ranges of both smaller and larger training resolutions in the main paper.
\begin{table}[htbp]
	\caption{Comparison of ResNet-18 baseline models (\textit{top}) and ResNet-18-based SAN (\textit{bottom}) with a range of medium training resolutions. The shaded numerical values have the same meaning as those in Table~\ref{tab:san-comparison} of the main paper.}
	\label{tab:resnet18-baseline-medium}
	\centering
	\resizebox{.8\linewidth}{!}{
		\begin{tabular}{c|ccccccccccccc}
			Train $\backslash$ Test & 384 & 320 & 256 & 224 & 208 & 192 & 176 & 160 & 144 & 128 & 112 & 96 & 64 \\
			\midrule[0.1em]
			320 & \cellcolor{gray}73.42 & \cellcolor{gray}72.56 & 70.03 & 67.44 & 64.92 & 63.23 & 59.47 & 56.07 & 51.17 & 45.89 & 38.64 & 30.27 & 11.74 \\
			256 & 72.75 & 72.68 & \cellcolor{gray}71.71 & 70.24 & 68.00 & 67.27 & 64.05 & 62.12 & 57.20 & 53.27 & 45.45 & 38.00 & 17.00 \\
			224 & 71.87 & 72.36 & 71.90 & \cellcolor{gray}70.97 & 69.14 & 68.72 & 66.20 & 64.68 & 60.36 & 56.85 & 50.17 & 42.55 & 20.40 \\
			192 & 69.95 & 71.07 & 71.71 & 71.25 & \cellcolor{gray}69.64 & \cellcolor{gray}69.75 & 67.69 & 67.04 & 63.23 & 60.47 & 54.21 & 47.73 & 25.10 \\
			160 & 66.95 & 69.05 & 70.72 & 70.70 & 69.50 & 70.14 & \cellcolor{gray}68.47 & \cellcolor{gray}68.48 & \cellcolor{gray}65.61 & \cellcolor{gray}63.80 & \cellcolor{gray}58.54 & \cellcolor{gray}53.78 & \cellcolor{gray}31.03 \\
		\end{tabular}
	}
	\label{tab:resnet18-medium}
	\centering
	\resizebox{.8\linewidth}{!}{
		\begin{tabular}{c|ccccccccccccc}
			Train $\backslash$ Test & 384 & 320 & 256 & 224 & 208 & 192 & 176 & 160 & 144 & 128 & 112 & 96 & 64 \\
			\midrule[0.1em]
			320 & \cellcolor{gray}74.23 & \cellcolor{gray}74.48 & 73.30 & 71.92 & 70.57 & 69.66 & 67.31 & 65.66 & 61.68 & 58.36 & 52.10 & 45.22 & 23.85 \\
			256 & 73.25 & 74.19 & \cellcolor{gray}73.96 & 73.10 & 72.01 & 71.55 & 69.53 & 68.26 & 65.22 & 62.18 & 56.65 & 50.34 & 28.15 \\
			224 & 72.03 & 73.44 & 73.88 & \cellcolor{gray}73.31 & 72.33 & 71.12 & 70.35 & 69.31 & 66.64 & 63.99 & 58.81 & 52.77 & 30.52 \\
			192 & 70.35 & 72.11 & 73.34 & 73.23 & \cellcolor{gray}72.33 & \cellcolor{gray}72.32 & 70.95 & 70.10 & 67.70 & 65.49 & 60.77 & 55.25 & 33.13 \\
			160 & 67.57 & 70.12 & 72.14 & 72.43 & 71.67 & 71.99 & \cellcolor{gray}70.96 & \cellcolor{gray}70.52 & \cellcolor{gray}68.25 & \cellcolor{gray}66.70 & \cellcolor{gray}62.36 & \cellcolor{gray}57.67 & \cellcolor{gray}36.18 \\
		\end{tabular}
	}
\end{table}

\section{More Dense Sampled Resolutions for Inference}

To further prove the applicability of our method to universally scalable input images, we sample the test resolutions in a more dense style. Suppose the training resolution range is $\mathbb{S} = \{S_1, S_2, S_3, S_4, S_5\} = \{224, 192, 160, 128, 96\}$ with the interval of 32, the test resolution range is set as $\mathbb{T} = \{S \pm \frac12 \times 32 \vert S \in \mathbb{S}\} = \{208, 176, 144, 112\}$ in the main paper. In addition, we conduct experiments on the test resolution range in a more dense arrangement as $\mathbb{T}^\prime = \{S \pm \frac14 \times 32 \vert S \in \mathbb{S}\} = \{216, 200, 184, 168, 152, 136, 120, 104\}$. We show the evaluation results of SAN on $\mathbb{T}^\prime$ with the proxy inference method in the bottom panel of Table~\ref{tab:resnet18-dense}. We also provide  baseline models individually trained on more dense input resolutions in the top panel of Table~\ref{tab:resnet18-dense}. It is observed that in the corresponding column, the shaded value (in the bottom sub-table) from our SAN achieves higher accuracy on these unseen resolutions than all evaluation results from the baseline models (in the top sub-table), demonstrating its versatility to a wide range of test resolutions. In other words, SAN is competent in processing input images well with arbitrary resolutions during inference.

\begin{table}[htbp]
	\caption{Comparison of ResNet-18 baseline models (\textit{top}) and ResNet-18-based-SAN (\textit{bottom}), with a more dense setting of test resolutions. The shaded results have the same meaning as those in Table~\ref{tab:san-comparison} of the main paper.}
	\centering
	\label{tab:resnet18-dense}
	\resizebox{.6\linewidth}{!}{
		\begin{tabular}{c|ccccccccccc}
			Train $\backslash$ Test & 216 & 200 & 184 & 168 & 152 & 136 & 120 & 104 \\
			\midrule[0.1em]
			224 & 70.40 & 68.28 & 67.69 & 64.55 & 63.18 & 57.60 & 54.20 & 45.26 \\
			208 & 70.11 & 69.77 & 68.12 & 66.81 & 63.77 & 61.44 & 55.67 & 50.27 \\
			192 & 70.88 & 69.05 & 69.04 & 66.52 & 65.75 & 60.83 & 58.41 & 49.98 \\
			176 & 70.56 & 70.50 & 69.35 & 68.44 & 66.25 & 64.14 & 59.55 & 54.95 \\
			160 & 70.43 & 68.94 & 69.70 & 67.13 & 67.50 & 63.49 & 61.89 & 54.46 \\
			144 & 69.89 & 69.84 & 69.47 & 68.84 & 67.61 & 66.16 & 62.71 & 59.08 \\
			128 & 68.55 & 66.97 & 68.78 & 66.66 & 67.91 & 64.49 & 64.68 & 57.83 \\
			112 & 66.37 & 66.63 & 67.37 & 67.44 & 67.16 & 66.49 & 64.46 & 62.26 \\
			96 & 59.24 & 57.66 & 62.56 & 60.64 & 64.58 & 61.92 & 64.63 & 60.18 \\
		\end{tabular}
	}
	\resizebox{.6\linewidth}{!}{
		\begin{tabular}{c|ccccccccccc}
			Train $\backslash$ Test & 216 & 200 & 184 & 168 & 152 & 136 & 120 & 104 \\
			\midrule[0.1em]
			224 & \cellcolor{gray}72.62 & 71.10 & 71.34 & 69.30 & 68.75 & 65.28 & 63.23 & 56.55 \\
			192 & 72.58 & \cellcolor{gray}71.33 & \cellcolor{gray}71.94 & 70.07 & 70.09 & 66.86 & 65.49 & 59.31  \\
			160 & 71.90 & 70.78 & 71.81 & \cellcolor{gray}70.14 & \cellcolor{gray}70.76 & 67.76 & 67.16 & 61.74 \\
			128 & 70.27 & 69.05 & 70.78 & 69.30 & 70.58 & \cellcolor{gray}67.83 & \cellcolor{gray}68.13 & 63.30 \\
			96 & 65.60 & 64.26 & 67.63 & 65.80 & 68.63 & 66.03 & 67.60 & \cellcolor{gray}63.34 \\
		\end{tabular}
	}
\end{table}

\section{Visualization of Privatized BN Layers}
\label{sec:visualization}

\begin{figure}[htbp]
	\begin{center}
		\includegraphics[width=\linewidth]{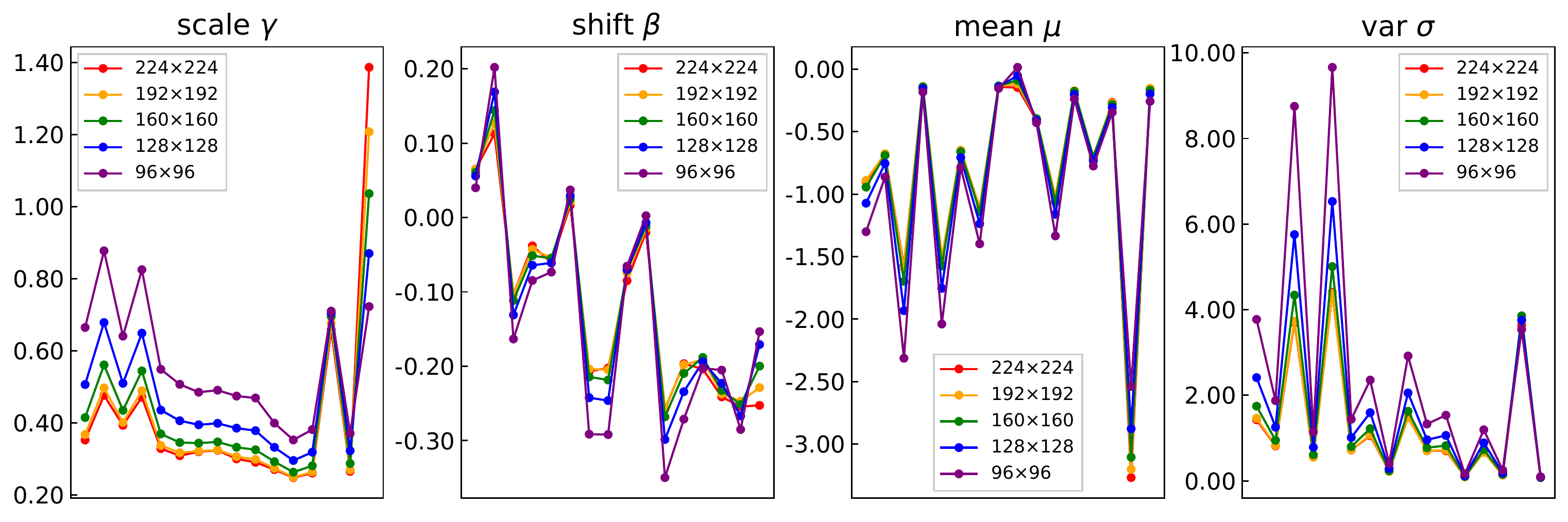}
	\end{center}
	\caption{Visualization of parameters and statistics of privatized BN layers in ResNet-18-based SAN. The x-axis represents the residual block index while the y-axis represents the mean value.}
	\label{fig:vsualization}
\end{figure}

Regarding privatized BN layers of ResNet-18-based SAN for each training resolution, we visualize their trainable parameters (scale and shift parameter $\gamma$ and $\beta$) and statistics (running mean $\mu$ and variance $\sigma$) in Fig.~\ref{fig:vsualization}. Specifically, we compute the mean values of these parameters across channels. There exist eight residual blocks in ResNet-18 and each residual block contains two BN layers. We plot the same type of BN parameters from all the sixteen layers in one sub-figure and make the observations that the mean values of BN parameters in each layer varies monotonically with respect to the training resolutions, \textit{i}.\textit{e}., following an either ascending or descending distribution with the training resolutions increasing. This phenomenon partially interprets the proper approximation of the proxy inference method and inspires the following data-free inference method.

\begin{table}[htbp]
	\caption{Accuracy of ResNet-18-based-SAN with data-free ideal inference and proxy inference respectively. The results of proxy inference are identical to those in Table~\ref{tab:san-comparison} of the main paper and Table~\ref{tab:resnet18-dense} of the supplementary materials.}
	\centering
	\label{tab:DF-ideal}
	\resizebox{\linewidth}{!}{
		\begin{tabular}{c|ccccccccccccccc}
			Inference $\backslash$ Resolution & 216 & 208 & 200 & 184 & 176 & 168 & 152 & 144 & 136 & 120 & 112 & 104 \\
			\midrule[0.1em]
			Data-Free Ideal & 72.66 & 72.08 & 71.32 & 72.00 & 71.15 & 70.20 & 70.78 & 69.35 & 67.91 & 68.11 & 65.95 & 63.38 \\
			Proxy & 72.62 & 72.10 & 71.33 & 71.94 & 71.04 & 70.14 & 70.76 & 69.35 & 67.83 & 68.13 & 65.78 & 63.34 \\
		\end{tabular}
	}
\end{table}

\section{Data-Free Ideal Inference via BN Interpolation}
In consideration of the monotonicity and continuity of BN parameters as visualized in Section~\ref{sec:visualization}, we come up with another solution to derive BN parameters for the ideal inference method. For an arbitrary test resolution $T$, we first search the nearest training resolutions located in its two sides, represented as $S_{floor}(T)$ and $S_{ceil}(T)$ where $S_{floor}(T) < S_{ceil}(T)$. Then we compute the BN parameters via a linear interpolation between these two groups of BN parameters, $\text{BN}^{(S_{floor}(T))}$ and $\text{BN}^{(S_{ceil}(T))}$. Specifically, the interpolation of BN is defined as
\begin{equation}
	\label{eqn:interpolation}
	\text{BN}^{(T)} = \frac{T - S_{floor}(T)}{S_{ceil}(T) - S_{floor}(T)}\text{BN}^{(S_{ceil}(T))} + \frac{S_{ceil}(T) - T}{S_{ceil}(T) - S_{floor}(T)}\text{BN}^{(S_{floor}(T))},
\end{equation}
where the parameters ($\gamma$, $\beta$, $\mu$ and $\sigma$) of BN are interpolated separately. The main network $\mathcal{N}^{(T)}$ could be parameterized with $\mathcal{M}(\varepsilon^{(T)})$, $\text{BN}^{(T)}$ and FC. It is noteworthy that this method does not depend on recalculating statistics of the training set anymore, so it is called \textit{data-free ideal inference}. The detailed inference process with this data-free BN computation method is depicted in Algorithm~\ref{alg:DF-ideal}. We report the performance of this method on the set of interpolated test resolutions $\mathbb{T} \cup \mathbb{T}^\prime$ and compare it to the proxy inference method (mainly used in our experiments) in Table~\ref{tab:DF-ideal}. We could see that data-free ideal inference is a more independent method compared with the original ideal inference while retaining the performance.

\section{Border and Round-off Effects}

\begin{figure}[htbp]
	\begin{center}
		\includegraphics[width=.5\linewidth]{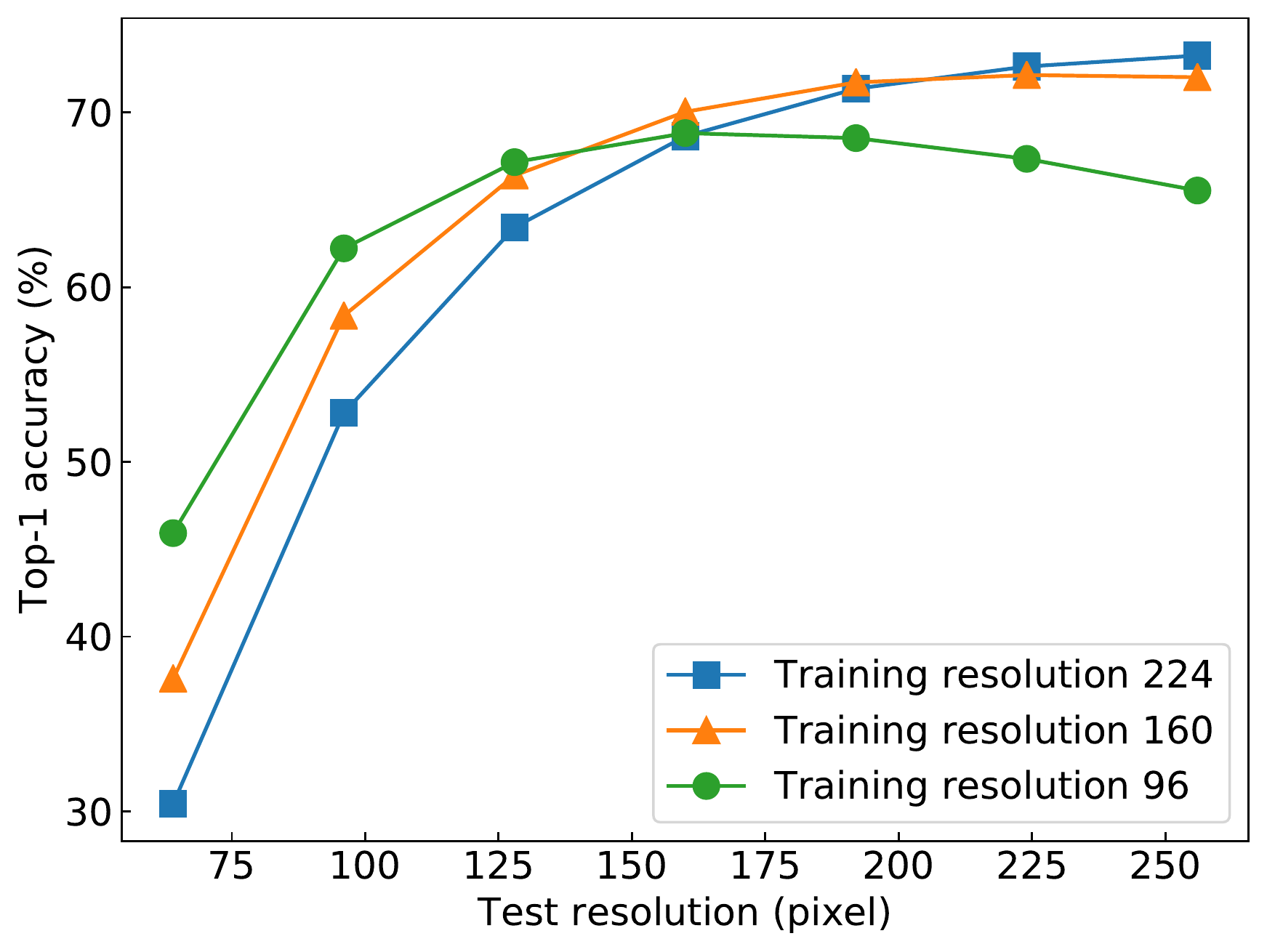}
	\end{center}
	\vskip -0.2in
	\caption{Validation accuracy curves of ResNet-18-based SAN evaluated on a spectrum of test resolutions which \textit{are} divisible by 32. The training resolutions are set to 224, 160 and 96, so test resolutions 256, 192, 128 and 64 are still interpolated or extrapolated operating points yet divisible by 32.}
	\label{fig:interpolation}
\end{figure}

It has drawn our attention that the curve of performance in Fig.~\ref{fig:envelop} of the main paper tends to fluctuate at certain test resolutions that are not divisible by 32 (modern neural networks usually have a down-sampling rate of $1/32$, hence these resolutions lead to discarded pixels of the intermediate feature maps), resembling the discovery of border and round-off effects discussed in Appendix of~\cite{touvron2019FixRes}. It is also noted that the aforementioned resolutions non-divisible by 32 are exactly the test resolutions in $\mathbb{T} \cup \mathbb{T}^\prime$ sandwiched by the training ones in $\mathbb{S} = \{S_1, S_2, S_3, S_4, S_5\} = \{224, 192, 160, 128, 96\}$. Our SAN model could achieve strong performance even at these interpolation points of training resolutions as demonstrated in Table~\ref{tab:DF-ideal}, which allows the evaluated images to be universally scalable. We further conduct experiments to prove by contradiction that the degradation phenomenon at these test resolutions does not arise from interpolation of training resolutions but essentially owing to the discrete nature of convolutional kernel sizes and strides. For the ResNet-18-based-SAN, we select three resolutions $\mathbb{S}=\{S_1, S_2, S_3\}=\{224,160,96\}$ for training and interpolate test resolutions which \textit{are} multiples of 32 during inference, shaping convex accuracy curves \textit{without} fluctuation in Fig.~\ref{fig:interpolation}.

\begin{table}[htbp]
	\caption{Comparison of ResNet-18 baseline models individually trained on the test resolutions and ResNet-18-based-SAN with proxy inference.}
	\centering
	\label{tab:round-off}
	\resizebox{\linewidth}{!}{
		\begin{tabular}{c|ccccccccccccccc}
			Method $\backslash$ Resolution & 216 & 208 & 200 & 184 & 176 & 168 & 152 & 144 & 136 & 120 & 112 & 104 \\
			\midrule[0.1em]
			baseline & 70.99 & 70.59 & 70.50 & 70.09 & 69.49 & 69.25 & 67.61 & 68.23 & 67.19 & 65.45 & 64.69 & 62.82 \\
			SAN & 72.62 & 72.10 & 71.33 & 71.94 & 71.04 & 70.14 & 70.76 & 69.35 & 67.83 & 68.13 & 65.78 & 63.34 \\
		\end{tabular}
	}
\end{table}

To alleviate the round-off effect for baseline models, we further individually train them on these test resolutions from scratch, establishing a stronger baseline for comparison. The results are compared with our SAN (with proxy inference) again in Table~\ref{tab:round-off}. Even compared to these better-performing baseline models that are pre-trained on the test resolutions, our SAN models still show superiority regarding the top-1 validation accuracy on ImageNet.

\section{Training Baseline Longer}
One may contradict that the training budget of SAN is much larger than a \textit{single} baseline model. However, this contradiction might overlook that although SAN training is based on more than one main network, it competes against \textit{as many baseline models as possible} after a single run of training. Therefore, we would like to argue that our comparison in the main paper follows a fair setting and a series of baseline models may even consume more computational resources than SAN in total. For instance, we simultaneously train on 5 resolutions for 150 epochs, then the training cost is roughly the same as separately training 5 baseline models on these resolutions, each for 150 epochs. In Fig.\ref{fig:envelop} and Table~\ref{tab:san-comparison} of the main paper, our single SAN is compared to 5 such models respectively at their theoretically best operating point (when test resolution meets the training one). Furthermore, if another unseen-sized image comes, SAN could directly perform the inference. But we may need to retrain a model on the new test resolution to obtain a near-optimal result for baseline, just like what the 5 previous baseline models do. 
\begin{table}[htbp]
	\caption{Accuracy of ResNet-18 baseline models on a spectrum of test resolutions when trained for 450 epochs, \textit{i}.\textit{e}., $3 \times 150$ epochs.}
	\label{tab:baseline}
	\vskip -0.1in
	\centering
	\resizebox{.75\linewidth}{!}{
		\begin{tabular}{c|ccccccccccc}
			Train $\backslash$ Test & 256 & 224 & 208 & 192 & 176 & 160 & 144 & 128 & 112 & 96 & 64 \\
			\midrule[0.1em]
			224 & \cellcolor{gray}73.55 & \cellcolor{gray}72.63 & 71.18 & 70.63 & 68.59 & 66.64 & 63.00 & 59.13 & 52.40 & 44.41 & 21.25 \\
			192 & 73.04 & 72.70 & \cellcolor{gray}71.84 & \cellcolor{gray}71.56 & 69.98 & 68.55 & 66.13 & 62.48 & 57.25 & 49.40 & 25.04 \\
			160 & 71.34 & 71.92 & 71.39 & 71.46 & \cellcolor{gray}70.37 & \cellcolor{gray}70.16 & 67.68 & 65.46 & 61.00 & 54.62 & 30.50 \\
			128 & 65.13 & 67.41 & 67.31 & 68.73 & 68.06 & 69.02 & \cellcolor{gray}67.72 & \cellcolor{gray}67.37 & 63.55 & 59.16 & 37.06 \\
			96 & 47.44 & 53.72 & 53.41 & 59.18 & 58.99 & 63.64 & 63.00 & 65.73 & \cellcolor{gray}63.82 & \cellcolor{gray}63.35 & \cellcolor{gray}46.82 \\
		\end{tabular}
	}
\end{table}
Now, the total training cost of baseline models has surpassed SAN. On the other hand, if one consists on considering \textit{only a single baseline model} for comparison, we train each baseline model for $3\times$ epochs (roughly the same wall-clock time as SAN training, sufficient for convergence), but only obtain marginal gains compared to the original baselines (cf. Table~\ref{tab:san-comparison} top-left in the main paper), as illustrated in Table~\ref{tab:baseline}. More importantly, they still perform poorly on other test resolutions with the enlarged training budget, due to possibly over-fitting to the training resolution.

%
%
\bibliographystyle{splncs04}
\bibliography{egbib}
\end{document}